\documentclass{article}

    \PassOptionsToPackage{numbers}{natbib}
  \usepackage[final]{neurips_2019}

\usepackage[utf8]{inputenc} % allow utf-8 input
\usepackage[T1]{fontenc}    % use 8-bit T1 fonts
\usepackage[colorlinks = true,
            linkcolor = red,]{hyperref}
\usepackage{url}            % simple URL typesetting
\usepackage{booktabs}       % professional-quality tables
\usepackage{amsfonts}       % blackboard math symbols
\usepackage{nicefrac}       % compact symbols for 1/2, etc.
\usepackage{microtype}      % microtypography

\usepackage{times}
\usepackage{epsfig}
\usepackage{graphicx}
\usepackage{amsmath}
\usepackage{amssymb}
\usepackage{xcolor}
\usepackage{adjustbox}
\usepackage{caption}
\usepackage{wrapfig}

\usepackage{amsmath}
\DeclareMathOperator*{\argmax}{arg\,max}
\DeclareMathOperator*{\argmin}{arg\,min}

\newcommand{\model}{DIB-R}
\newcommand{\vertexdata}{attribute}
\newcommand{\vertexdatas}{attributes}

\title{Learning to Predict 3D Objects with an Interpolation-based Differentiable Renderer}

\author{%
Wenzheng Chen$^{1,2,3}$  \And Jun Gao$^{1,2,3,}$\thanks{authors contributed equally}  \And Huan Ling$^{1,2,3,*}$  \And Edward J. Smith$^{1,4,*}$ \\
\And Jaakko Lehtinen$^{1,5}$ \quad \textbf{Alec Jacobson$^{2}$} \quad \textbf{Sanja Fidler$^{1,2,3}$} \vspace{6pt}\\
\small{NVIDIA\textsuperscript{1} \quad University of Toronto\textsuperscript{2} \quad Vector Institute\textsuperscript{3} \quad McGill University\textsuperscript{4} \quad Aalto University\textsuperscript{5}} \vspace{3pt}\\
\texttt{\scriptsize \{wenzchen, huling, jung, esmith, jlehtinen, sfidler\}@nvidia.com,
jacobson@cs.toronto.edu}
}
\begin{document}

\maketitle
% \vspace{-5.6mm}
\begin{abstract}
%\alec{You might want to borrow from Derek's ICLR paper's motivation a bit and add a sentence here or in the introduction about how most ML models operate on pixel-images, but real images are formed by light interacting with geometry. Understanding the process of going from light+geometry to images (``Rendering'') is important for ML to make progress.}
 % 1 render is important but non-differentiable
%Rendering pipelines provide an important link between 3D models and 2D images, by projecting geometry and simulating its natural interaction with light. 
%Most Machine Learning (ML) models operate on images which are 2D projections that have been formed by 3D geometry interacting with light through the process called rendering. %\alec{this is burying the point. most ML vision models operate on images, but they \emph{ignore} the fact that images are formed via light+geometry} Enabling ML models to understand the process of image formation might be key for generalization. 
%Images are 2D projections that are formed by 3D geometry interacting with light through the process called rendering. 
Many machine learning models operate on images, but ignore the fact that images are 2D projections formed by 3D geometry interacting with light, in a process called rendering. Enabling ML models to understand image formation might be key for generalization. 
However, due to an essential rasterization step involving discrete assignment operations, rendering pipelines are non-differentiable and thus largely inaccessible to gradient-based ML techniques. In this paper, we present {\emph \model}, a differentiable rendering framework which allows gradients to be analytically computed for all pixels in an image. Key to our approach is to view foreground rasterization as a weighted interpolation of local properties and background rasterization as a distance-based aggregation of global geometry. Our approach allows for accurate optimization over vertex positions, colors, normals, light directions and texture coordinates through a variety of lighting models. We showcase our approach in two ML applications: single-image 3D object prediction, and 3D textured object generation, both trained using exclusively using 2D supervision. Our project website is: \href{https://nv-tlabs.github.io/DIB-R/}{https://nv-tlabs.github.io/DIB-R/}

\end{abstract}

% unsupervised 2d 
% first
\vspace{-17pt}
\section{Introduction}
\vspace{-10pt}
 % want to mention the verious lighting models we provide 
 % less fluff, more stuff 
3D visual perception contributes invaluable information when understanding and interacting with the real world. However, the raw sensory input to both human and machine visual processing streams are 2D projections (images), formed by the complex interactions of 3D geometry with light. Enabling machine learning models to understand the image formation process could facilitate disentanglement of geometry from the lighting effects, which is key in achieving invariance and robustness.
  
  %: Do we really need to see many images of the same object in different lighting conditions when we could imagine them? 
%
%This sensory process fuses shape, color, texture, lighting, and many more informative properties of the 3D environment into an easily digestible 2D representation. \alec{This sentence is your main point. You don't need to make claims about what sensory input is the ``most important'' for human interaction/learning. Readers will already agree that vision is important. We want to stress that vision is not just receiving pixel images, but rather measuring lights' interaction with 3D shapes.}
%
%Most forms of human learning depends upon the visual feedback loop that this projection allows and it is quite daunting to imagine developing many skills without it.

The process of generating a 2D image from a 3D model is called \emph{rendering}. Rendering is a well understood process in graphics with different algorithms developed over the years. %These fall into different categories based on how the light transport is being modeled. \emph{Rasterization}-based techniques are the fastest as they only geometrically project 3D objects onto the image plane but do not model more advanced lighting effects such as shadows and indirect light. \emph{Ray casting} and \emph{ray tracing} algorithms are able to handle these effects via more complex optical simulation. 
Making these pipelines amenable to deep learning requires us to differentiate through them.

In~\cite{li2018differentiable}, the authors introduced a differentiable ray tracer which builds on Monte Carlo ray tracing, and can thus deal with secondary lighting effects such as shadows and indirect light. Most of the existing work focuses on rasterization-based renderers, which, while simpler in nature as they geometrically project 3D objects onto the image plane and cannot support more advanced lighting effects, have been demonstrated to work well in a variety of ML applications such as single-image 3D prediction~\cite{loper2014opendr,NMR,liu2019soft, liu2019soft_v2}. Here, we follow this line of work.

Existing rasterization-based approaches typically compute approximate gradients~\cite{loper2014opendr,NMR} which impacts performance. Furthermore, current differentiable rasterizertion methods fail to support differentiation with respect to many informative scene properties, such as textures and lighting, leading to low fidelity rendering, and less informative learning signals~\cite{NMR, liu2019soft, liu2019soft_v2} . 

In this paper, we present {\model }, an approach to differentiable rendering, which, by viewing rasterization as a combination of local interpolation and global aggregation, allows for the gradients of this process to be computed analytically over the entire image. When performing rasterization of a foreground pixel, similar to~\cite{genova2018unsupervised}, we define its value as a weighted interpolation of the relevant vertex attributes of the foreground face which encloses it. To better capture shape and occlusion information in learning settings we define the rasterization of background pixels through a distance-based aggregation of global face information. With this definition the gradients of produced images can be passed back through a variety of vertex shaders, and computed with respect to all influencing vertex attributes such as positions, colors, texture, light; as well as camera positions. Our differentiable rasterization's design further permits the inclusion of several well known lighting models.

%With this advancement, for the first time in this setting learning systems can develop an understanding of the interplay between geometry and lighting, and disentangle their contribution to rendered images. By interpolating over directions of light, normal and eye, we support the Lambertian reflectance~\cite{lambert1760photometria} where local illumination is decided by normal and light and Phong shading~\cite{Phong1975} which \ed{fill in with info on phong  model} is supported by  \ed{fill in with the properties we need}. Through the incorporation of  \ed{fill in with the properties we need}, we also support Spherical Harmonic lighting, where \ed{fill in with info on SH model}.  %In contrast to previous approaches for differentiable rendering in the context of machine learning, our method can easily digest full scene information, to produce realistic images across a wide variety of mesh properties and environment effect. 
%\alec{Is it really new that we support these? Or is it new that we show that the parameters of these lighting models can be effectively learned from images?} 
%\ed{we are the first diff rasterizer to support this so I think we can claim right? } 
%\wz{Other methods, like OpenDR and Paparazi, support SH. But we are the first try to use ML to distangle texture from light}

We wrap our {\model } around a simple neural network in which the properties of an initial polygon sphere are predicted with respect to some conditioning input. We showcase this framework in a number of challenging machine learning applications focusing on 3D shape and texture recovery, across which we achieve both numerical and visual state-of-the art results. 
%First, we apply it to the task of single-image 3D object reconstruction using only 2D supervision. On this task we achieve state-of-the-art results both visually and numerically. 
%Second, through various ablation studies we demonstrate that our model is able to learn to not only reconstruct accurate shape, but also textures and lighting, with rendered predicted objects almost indistinguishable from the ground truth test images. Finally, with our rendering pipeline acting as a generator in an adversarial framework, we learn distributions of 3D textured objects found in the ShapeNet dataset~\cite{ShapeNet}, using exclusively 2D supervision, in the form of RGB images.
%\alec{Do we need to define/explain what we mean by ``exclusively 2D supervision'' ?}

\vspace{-7pt}
\section{Related work}
\vspace{-5pt}
\paragraph{Differentiable Rasterization:} 
%\vspace{-5pt}
% talk about differences and similarities of difference diff renderers: 
% nural mesh renderer
% soft rasterizers
% papparazi
% Differentiable Monte Carlo Ray Tracing through Edge Samplin
% OpenDR: An Approximate Differentiable Renderer
% RenderNet: A deep convolutional network fordifferentiable rendering from 3D shapes

 OpenDR~\cite{loper2014opendr}, the first in the series of differentiable rasterization-based renderers, approximates gradients with respect to pixel positions using first-order Taylor approximation, and uses automatic differentiation to back-propagate through the user-specified forward rendering program. In this approach, gradients are non-zero only in a small band around the edges of the mesh faces, which is bound to affect performance.~\cite{NMR} hand-designs an approximate gradient definition for the movement of faces across image pixels. %, to allow object silhouettes to be rendered for object generation tasks, without 3D supervision.
 The use of approximated gradients, and lack of full color information results in noisy 3D predictions, without concave surface features. To analytically compute gradients, Paparazzi~\cite{liu2018paparazzi} and~\cite{liuadvgeo2018}, propose to back-propagate the image gradients to the face normals, and then pass them to vertex positions via chain rule. However, their gradient computation is limited to a particular lighting model (Spherical Harmonics), and the use of face normals further prevents their approach to be applied to smooth shading. 
\cite{petersen2019pix2vex} designs a $C^{\infty}$ smooth differetiable renderer for estimating 3D geometry, while neglecting lighting and texture. \cite{szabo2018unsupervised} supports per-vertex color and approximates the gradient near boundary with blurring, which produces wired effects and can not cover the full image. \cite{insafutdinov2018unsupervised} focus on rendering of point cloud and adopts a differentiable reprojection loss to constrain the distribution of predited point clouds, which loses point connectivity and cannot handle texture and lighting.

SoftRas-Mesh recently proposed in~\cite{liu2019soft} introduces a probabilistic formulation of rasterization, where each pixel is softly assigned to \emph{all} faces of the mesh. While inducing a higher computational cost, this clever trick allows gradients to be computed analytically. Parallel to our work, SoftRas-Color~\cite{liu2019soft_v2} extended this framework to incorporate vertex colors and support texture and lighting theoretically. % However, this approach relies on a particularly designed color imaging formation model, making extensions to other vertex attributes such as texture coordinates or normals, non-trivial.
However, in~\cite{liu2019soft_v2} each pixel would be influenced by all the faces and thus might have blurry problem.
The key difference between the parallel work of ~\cite{liu2019soft_v2} and ours is that, similarly to \cite{genova2018unsupervised}, % we compute analytic gradients of foreground pixels by viewing rasterization as interpolation of \emph{local} mesh properties. This more naturally supports optimization with respect to all vertex attributes, and additionally enables the extension of our pipeline to a variety of different lighting models. which~\cite{liu2019soft_v2} cannot support.
we specify each foreground pixel to the most front face and compute analytic gradients of foreground pixels by viewing rasterization as interpolation of \emph{local} mesh properties. This allows our rendering effect the same as OpenGL pipeline and naturally supports optimization with respect to all vertex attributes, and additionally enables the extension of our pipeline to a variety of different lighting models.
In contrast to~\cite{genova2018unsupervised}, which also uses an interpolation-based approach, but applied to the entire image, our rasterization module allows for soft assignment of background pixels through an aggregation of global features. % \wz{I delete the last sentence since it is too long} When learning, this enables a better understanding of potential updates to geometry, and information provided by occluded faces. In addition, our pipeline allows for rendering and learning over more advanced mesh properties such a textures and camera positions and different lighting models from graphics such as Lambertian and Spherical Harmonics. 

%\alec{What about MeshCNN or voxel GAN papers? Should at least cite them and state that they train using 3D supervision? I guess that's what this iffalse paragraph is below. Does NeurIPS have a page limit? Is that why this is left out?}

\vspace{-3pt}
\paragraph{Adverserial 3D Object Generation:}

Generation of 3D shapes through deep learning has been approached using a Generative Adverserial Network (GAN)~\cite{goodfellow2014generative} in
a plethora of work
%the adoption of adversarial regularizers to encourage the prediction of more likely objects
~\cite{yang20173d, achlioptas2017learning, wu2018learning, 3DIWGAN}. While these approaches require full 3D supervision, differentiable rendering frameworks allow learning 3D object distributions using only 2D supervision~\cite{henderson2018learning}. We showcase our model in the same application, where we are the first to learn a generator for both shape and texture. %\ed{we are the first GAN on meshes  as well?}% using only 2D supervision.

%In contrast to these approaches which require full 3D supervision, our differentaible rendering framework allows us to learn 3D object distributions  using only 2D supervision. While~\cite{henderson2018learning} previously demonstrated this was possible, as our differentiable rasterizer also allows texture predictions, we boast the first 3D GAN system able to learn both shape and texture using only 2D supervision.  

\vspace{-5pt}
\section{{\model }: Differentiable Interpolation-based  Renderer}
\vspace{-5pt}
In this section, we introduce our {\model }. Treating foreground rasterization as an interpolation of vertex attributes allows realistic images to be produced, whose gradients can be fully back-propagated through all predicted vertex \vertexdatas, while defining background rasterization as an aggregation of global information during learning allows for better understanding of shape and occlusion. 

% First, discuss render pipeline in OpenGL
\vspace{-7pt}
\subsection{Rendering Pipeline}
\vspace{-5pt}
Many popular rendering APIs, such as OpenGL~\cite{woo1999opengl} and DirectX3D~\cite{luna2012introduction}, decompose the process of rendering 3D scenes into a set of sequential user-defined programs, referred to as \emph{shaders}. While there exist many different shader types, the vertex, rasterization, and fragment shaders the three most important steps for establishing a complete rendering pipeline. When rendering an image from a 3D polygon mesh, first, the vertex shader projects each 3D vertex in the scene onto the defined 2D image plane. Rasterization is then used to determine which pixels are covered and in what manner, by the primitives these vertices define. Finally, the fragment shader computes how each pixel is colored by the primitives which cover it.  

%While each of these shaders is essential for any colored rendering, for the purposes of designing a differentiable rending pipeline, rasterization must take the highest consideration due to the inherently non-differentiable operations which it requires. In contrast, 
The vertex and fragment shaders can easily be defined such that they are entirely differentiable. By projecting 3D points onto the 2D image plane by multiplying with the corresponding 3D model, view and projection matrices, the vertex shader operation is directly differentiable. In the fragment shader, pixel colors are decided by a combination of local properties including assigned vertex colors, textures, material properties, and lighting. While the processes through which this information are combined can vary with respect to the chosen rendering model, in most cases this can be accomplished through the application of fully differentiable arithmetic operations. All that remains for our rendering pipeline is the rasterization shader, which presents the main challenge, due to the inherently non-differentiable operations which it requires. In the following section we describe our method for rasterizing scenes such that the derivatives of this operation can be analytically determined. % We note that our approach in its current form does not support handling occlusion. While pixels could be covered by multiple faces with different depth, our approach only considers the most front face and is thus limited in the scenes it can handle. Accounting for occlusion is future work. % \alec{This is not clear enough. Is occlusion handled per-iteration via rasterization and keeping only the closest fragment (z-buffer) as Paparazzi does? Or is every triangle that lands on a pixel considered to be rendered? If it's the later is the color taken as an average?} \wz{the former}
% \alec{Still not clear. This seems to contradict Eq. 6} \wz{put the explanation at the end}

\vspace{-6pt}
\subsection{Differentiable Rasterization}
\vspace{-5pt}

\begin{wrapfigure}{R}{0.25\textwidth} 
\centering
\vspace{-14pt}
\includegraphics[width=0.25\textwidth]{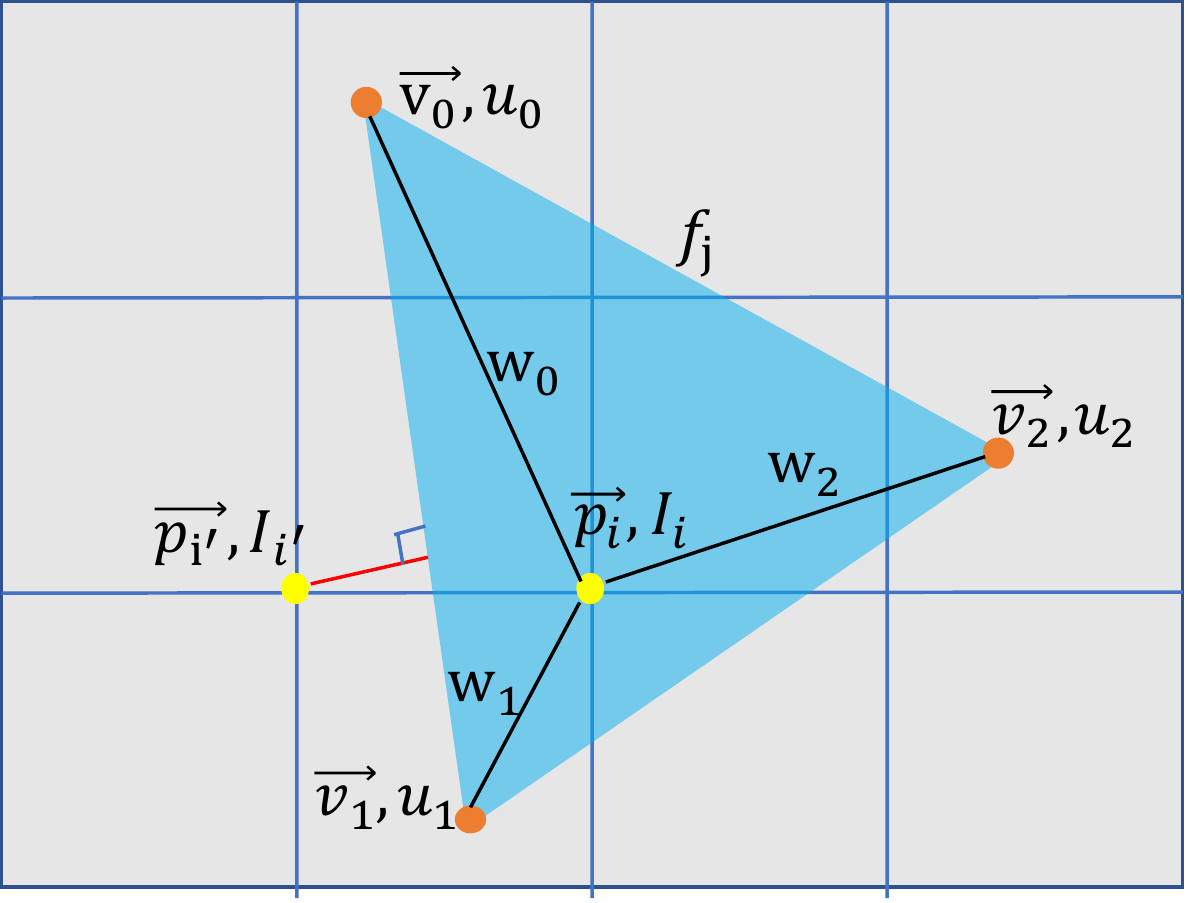}
 \vspace{-15pt}
 \label{fig:demonstrate}
 \caption{\footnotesize Illustration of our Differentiable Rasterization.}
\end{wrapfigure}
% show rasterization is differentiable
% need to revise veriable names in this section and reconsider interpolation formula
Consider first only the \textbf{foreground pixels} that are covered by one or more faces. Here, in contrast to standard rendering, where a pixel's value is assigned from the closest face that covers it, we treat foreground rasterization as an interpolation of vertex \vertexdatas  \cite{genova2018unsupervised}. For every foreground pixel we perform a z-buffering test~\cite{greene1993hierarchical}, and assign it to the closest covering face. Each pixel is influenced exclusively by this face. Shown in Fig.~1, a pixel at position $\vec{p}_i$ is covered by face $f_j$ with three vertices $\vec{v}_{0}$, $\vec{v}_{1}$, $\vec{v}_{2}$, and each vertex has its own \vertexdatas: $u_{0}$, $u_{1}$, $u_{2}$, respectively. $\vec{p}_i$ and  $\vec{v}_i$ are 2D coordinates on the image plane while $u_i$ are  scalars. We compute the value of this pixel, $I_i$, using barycentric interpolation % \alec{this citation doesn't make sense and is unnecessary}~\cite{berrut2004barycentric} 
of the face's vertex \vertexdatas:
\begin{equation}
    I_i = w_0 u_{0} + w_1 u_{1} + w_2 u_{2}, 
\label{equation:interpolation}
\end{equation}
where weights $w_0$, $w_1$ and $w_2$ are calculated over the vertex and pixel positions using a differentiable functions $\Omega_k$ (provided in Appendix):\\[-2mm] 
\begin{equation}
    w_k = \Omega_k(\vec{v}_{0}, \vec{v}_{1},\vec{v}_{2}, \vec{p}_i), \quad k=0, 1, 2.
\label{equation:wfunc}
\end{equation}
While barycentric interpolation has been widely used in OpenGL pipeline. Here, we derive the differentiable reformulation. With this approach, it is easy to back-propagate gradients from a loss function $L$, defined on the output image, through pixel value $I_i$ to vertex \vertexdatas {} $u_{k}$ via chain rule:\\[-2mm]
\begin{equation}
  \frac{\partial I_i}{\partial u_k} = w_k, \quad
  \frac{\partial I_i}{\partial \vec{v}_{k}} = \sum_{m=0}^{2} \frac{\partial I_i}{\partial w_m} \frac{\partial \Omega_m}{\partial \vec{v}_{k}},
\label{equation:diff2}
\end{equation}
\begin{equation}
  \frac{\partial L}{\partial u_k} = \sum_{i=1}^N \frac{\partial L}{\partial I_i} \frac{\partial I_i}{\partial u_k}
  ,\quad 
  \frac{\partial L}{\partial \vec{v}_k} = \sum_{i=1}^N \frac{\partial L}{\partial I_i} \frac{\partial I_i}{\partial \vec{v}_k},
\label{equation:diff3}
\end{equation}
% we need to mention the 
where $N$ is the number of pixels covered by the face. Now consider pixels which no faces cover, which we refer to as \textbf{background pixels}. Notice that in the formulation above, the gradients from background pixels cannot back-propagate to any mesh \vertexdatas. However, the background pixels provide a strong constraint on the 3D shape, and thus the gradient from them provide a useful signal when learning geometry. Take, for example, pixel $p_{i'}$ at position $\vec{p}_{i'}$ which lies outside of face $f_j$, in Fig~1. We want this pixel to still provide a useful learning signal. In addition, information from occluded faces an entirely ignored despite their potential future influence.

Inspired by the silhouette rasterizetion of~\cite{liu2019soft}, we define a distance-related probability $A^j_{i'}$, that softly assigns face $f_j$ to pixel $p_{i'}$ as:
%the probability that face $f_j$ influences pixel $p_i$, as: 
\begin{eqnarray}
	{A^j_{i'}} = \exp(-\frac{d({p_{i'}}, f_j)}{\delta}),
	\label{equation:diff4}
\end{eqnarray}
where $d({p_{i'}}, f_j)$ is the distance function from pixel $p_{i'}$ to face $f_j$ in the projected 2D space, and $\delta$ is a hyper-parameter that controls the smoothness of the probability (details provided in Appendix). We then combine the probabilistic influence of all faces on a particular pixel in the following way: \\[-5mm]
\begin{eqnarray}
	{A_{i'}} = 1 - \prod_{j=1}^n (1-{A^j_{i'}}).
	\label{equation:diff5}
\end{eqnarray}
where $n$ is the number of all the faces. The combination of all $A_{i'}$ into their respective pixel positions makes up our alpha channel prediction. With definition, any background pixel can pass its gradients back to positions of all the faces (including those ignored in the foreground pixels due to occlusion) with influence proportional to the distance between them in alpha channel.
%
% \alec{This is confusing. Which is used? This or the z-buffer?}\wz{we spit pixels into foreground pixels and back ground pixels, where foreground pixels we calculate gradients based on the z-buffer(each pixel is just related to one most front face that cover it). For background pixels they will be influenced by all the faces, and thus we use the formula rather than z-buffer.}
%
As all foreground pixels have a minimum distance to some face of $0$, they must receive an alpha value of 1, and so gradients will only be passed back through the colour channels as defined above.

In summary, foreground pixels, which are covered by a specific face, back-propagate gradients though interpolation while background pixels, which are not covered by any face, softly back propagate gradients to all the faces based on distance. In this way, we can analytically determine the gradients for all aspects of our rasterization process and have achieved a fully differentiable rendering pipeline. 

% \ed{I removed a paragraph here, bring is back if absolutely nessecary, though it serves no purpose}
%We note that our approach in its current form does not explicitly handle occlusion. Even though foreground pixels could be covered by multiple faces with different depth, due to z-buffering they are only influenced by the closest faces. In contrast, background pixels are related to all faces(Eq. ~\ref{equation:diff5}), however, due to distance attenuation they mostly affect faces near the outer boundary. Accounting for occlusion in an explicit way is future work. 

% For a more detailed derivation please refer to the supplementary.

% We believe our propability is a better simulation of silhouette than \cite{}. We show the difference of our propobability map and propobabilitu map in \cite{}. \cite{} use $Sigmoid$ function, which leads dard lines in face edge area. Instead, we simply use exp function and only process pixels outside faces, makes our propobability more xx adnxx.
\vspace{-5pt}
\subsection{Rendering Models}
\vspace{-5pt}
\label{subsec:rendermodels}
% The key advantage of our differentiable rasterization technique it that it can be embedded into a variety of advanced rendering models. In the following section we outline the vertex \vertexdatas the rasterizer can interpolate over and then back-propagate through, and the lighting models which the support of these \vertexdatas allows. A complete overview of this information is shown in Table ~\ref{tb:oglmodel}.

In Equation~\ref{equation:interpolation}, we define pixel values $I_i$ by the interpolation of abstract vertex {\vertexdatas} $u_0$, $u_1$ and $u_2$. As our renderer expects a mesh input, vertex position is naturally one such \vertexdata, but we also simultaneously support a large array of other vertex {\vertexdatas } as shown in the Appendix. %Table~\ref{tb:oglmodel}, Column 1. 
In the following section we outline the vertex {\vertexdatas } the rasterization can interpolate over and then back-propagate through, and the rendering models which the support of these {\vertexdatas } allows. A complete overview of this information is shown in Appendix. %Table ~\ref{tb:oglmodel}.

\vspace{-5pt}
\subsubsection{Basic Models}
\vspace{-5pt}
Our \model {} supports basic rendering models where we draw the image directly with either vertex colors or texture. To define the basic colours of the mesh we support vertex {\vertexdatas} as either vertex colour or u,v coordinates over a learned or predefined texture map. Pixel values are determined through bi-linear interpolation of the vertex colours, or projected texture coordinates, respectively.
%or can be enhanced with the advanced lighting systems described in the next sub-section. 
%To enable these lighting effects
%we can interpolate over further mesh properties such as vertex normals and lighting direction  and learn extrinsic scene parameters such as camera position, and camera orientation. 
% We can further interpolate over more rendering properties such as vertex normals  and learn extrinsic scene parameters such as camera position, and camera orientation.  
\vspace{-5pt}
\subsubsection{Lighting Models}
\vspace{-5pt}
We also support 3 different local illumination models: Phong~\cite{Phong1975}, Lambertian~\cite{lambert1760photometria} and Spherical Harmonics~\cite{spheircalharmonic}, where the lighting effect is related to normal, light and eye directions.

To unify all the different lighting models, we decompose image color, $I$, into a combination of mesh color $I_c$ and lighting factors $I_l$ and  $I_s$: 
\begin{equation}
I = I_l I_c + I_s. 
\end{equation}
$I_c$ denotes the interpolated vertex colour or texture map values extracted directly from the vertex \vertexdatas {} without any lighting effect, $I_l$ and $I_s$ donate the lighting factors decided by specific lighting model chosen, where $I_l$ will be merged with mesh colour and $I_s$ is additional lighting effect that does not rely on $I_c$. %such as the secular effects required in the Phong Model~\cite{}. 
We first interpolate light-related {\vertexdatas} such as normals, light or eye directions in rasterization, then apply different lighting models in our fragment shader.

\textbf{Phong and Lambertian Models:}
In the Phong Model, image colour $I$ is decided by vertex normals, light directions, eye directions and material properties through the following equations: 
\begin{equation}
I_l = k_d (\vec{L} \cdot \vec{N}), \quad   \quad I_s = k_s (\vec{R} \cdot \vec{V})^{\alpha},
\end{equation}
where, $k_d$, $k_s$ and $\alpha$ are: diffuse reflection, specular reflection, and shininess constants. $\vec{L}$, $\vec{N}$, $\vec{V}$ and $\vec{R}$ are directions of light, normal, eye and reflectance, respectively, which are all interpolated vertex \vertexdatas. This results in the following definition for image colour under the Phong model:
\begin{equation}
I_{Phong} = I_c k_d (\vec{L} \cdot \vec{N})  + k_s (\vec{R} \cdot \vec{V})^{\alpha} . 
\end{equation}
As a slight simplification of full Phong shading we do not adopt ambient light and set light colour at a constant value of $1$. The Lambertian model can be viewed as a further simplification of the Phong Model, where we only consider diffuse reflection, and set $I_s$ as zero: 
\begin{equation}
I_{Lambertian} = I_c k_d (\vec{L} \cdot \vec{N}). 
\end{equation}

\textbf{Spherical Harmonic Model:}
Here,   $I_l$ is determined by normals while $I_s$ is set to $0$: 
\vspace{-5pt}
\begin{equation}
I_{Spherical Harmonic} = I_c  \sum_{l=0}^{n-1}\sum_{m=-l}^{l}w_l^m Y_l^m(\vec{N}),
\vspace{-5pt}
\end{equation}
where $Y_m^l$ is an orthonormal basis for spherical functions analogous to a Fourier series where $l$ is the frequency and 
% denoted with the subscript with l as the frequency and that there are 2l + 1 functions per frequency, denoted by superscripts m
% between −l to l inclusively
$w_l^m$ is the corresponding coefficient for the specific basis. To be specific, we set $l$ as 3 and thus predict 9 coefficients in total. By adjusting different $w_l^m$, different lighting effects are simulated. For more details please refer to \cite{spheircalharmonic}, Section ~2.

%\subsection{Optimization}

\begin{figure}
% \vspace{-1mm}
    \centering
      \begin{minipage}{0.72\linewidth}
    \begin{minipage}[t] {0.041\textwidth}
    \rotatebox{90}{\scriptsize $\ \ $initial}
    \rotatebox{90}{\scriptsize $\ \ $guess}
    \end{minipage}
    \begin{minipage}[t] {0.09\textwidth}
    \centering
    \includegraphics[width=\linewidth]{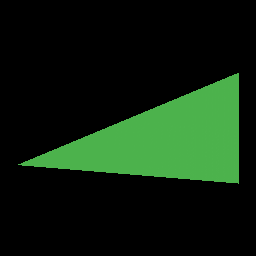}
    \end{minipage}
    \begin{minipage}[t] {0.09\textwidth}
    \centering
    \includegraphics[width=\linewidth]{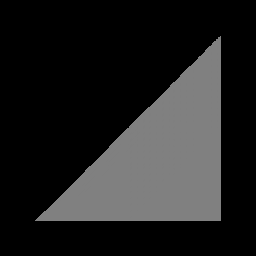}
    \end{minipage}
    \begin{minipage}[t] {0.09\textwidth}
    \centering
    \includegraphics[width=\linewidth]{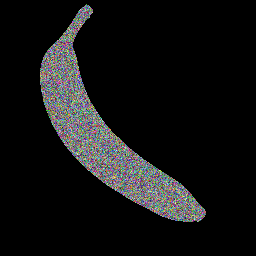}
    \end{minipage}
    \begin{minipage}[t] {0.09\textwidth}
    \centering
    \includegraphics[width=\linewidth]{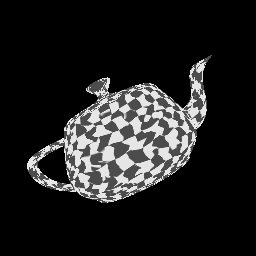}
    \end{minipage}
    \begin{minipage}[t] {0.09\textwidth}
    \centering
    \includegraphics[width=\linewidth]{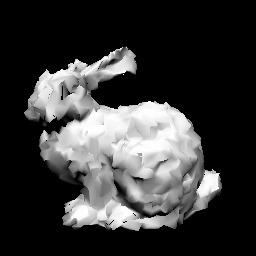}
    \end{minipage}
    \begin{minipage}[t] {0.09\textwidth}
    \centering
    \includegraphics[width=\linewidth]{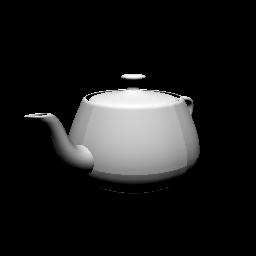}
    \end{minipage}
    \begin{minipage}[t] {0.09\textwidth}
    \centering
    \includegraphics[width=\linewidth]{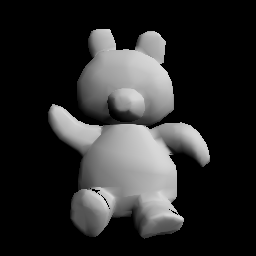}
    \end{minipage}
    \begin{minipage}[t] {0.09\textwidth}
    \centering
    \includegraphics[width=\linewidth]{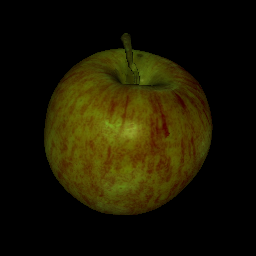}
    \end{minipage}
  %      \begin{minipage}[t] {0.09\textwidth}
  %  \centering
  %  \includegraphics[width=\linewidth]{figures/optim2/shininessini.png}
  %  \end{minipage}
    \\
    % \includegraphics[width=0.13\linewidth]{figures/optim/vc2ini.png}
    % \includegraphics[width=0.13\linewidth]{figures/optim/texini.png}
    % \includegraphics[width=0.13\linewidth]{figures/optim/uv3ini.png}
    % \includegraphics[width=0.13\linewidth]{figures/optim/ver2ini.png}
    % \includegraphics[width=0.13\linewidth]{figures/optim/cam2ini.png}
    % \includegraphics[width=0.13\linewidth]{figures/optim/li2ini.png}
    % \includegraphics[width=0.13\linewidth]{figures/optim/mat2ini.png}
    % \hspace{1pt}
        \begin{minipage}[t] {0.04\textwidth}
    \rotatebox{90}{\scriptsize $\ \  $target}\hspace{-0.8mm}
    \rotatebox{90}{\scriptsize $\ \ $img.}
    \end{minipage}
    \begin{minipage}[t] {0.09\textwidth}
    \centering
    \includegraphics[width=\linewidth]{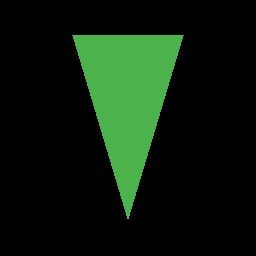}
    \end{minipage}
    \begin{minipage}[t] {0.09\textwidth}
    \centering
    \includegraphics[width=\linewidth]{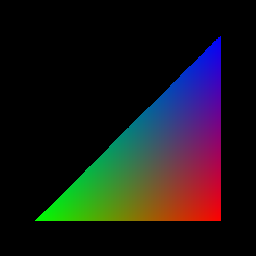}
    \end{minipage}
    \begin{minipage}[t] {0.09\textwidth}
    \centering
    \includegraphics[width=\linewidth]{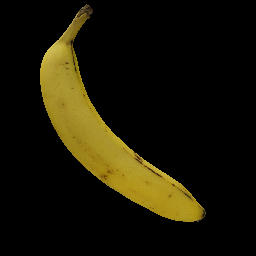}
    \end{minipage}
    \begin{minipage}[t] {0.09\textwidth}
    \centering
    \includegraphics[width=\linewidth]{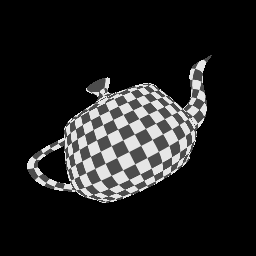}
    \end{minipage}
    \begin{minipage}[t] {0.09\textwidth}
    \centering
    \includegraphics[width=\linewidth]{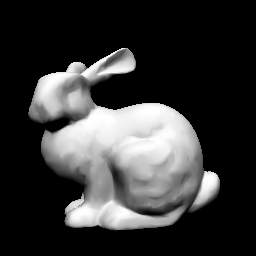}
    \end{minipage}
    \begin{minipage}[t] {0.09\textwidth}
    \centering
    \includegraphics[width=\linewidth]{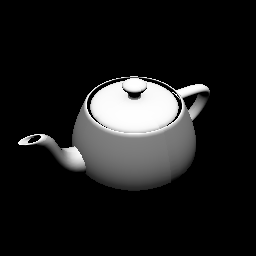}
    \end{minipage}
    \begin{minipage}[t] {0.09\textwidth}
    \centering
    \includegraphics[width=\linewidth]{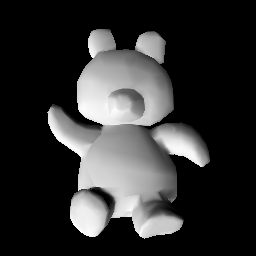}
    \end{minipage}
    \begin{minipage}[t] {0.09\textwidth}
    \centering
    \includegraphics[width=\linewidth]{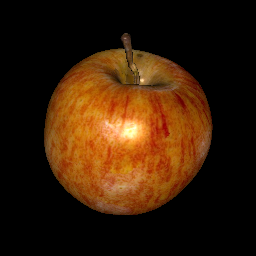}
    \end{minipage}
  %      \begin{minipage}[t] {0.09\textwidth}
  %  \centering
  %  \includegraphics[width=\linewidth]{figures/optim2/shininesstarget.png}
   % \end{minipage}
        \\
    \hspace{-0.5mm}
    \begin{minipage}[t] {0.041\textwidth}
    \rotatebox{90}{\scriptsize $\ \  $optim.}\hspace{-0.5mm}
    \rotatebox{90}{\scriptsize $\ \ $result}
    \end{minipage}
    \begin{minipage}[t] {0.09\textwidth}
    \centering
    \includegraphics[width=\linewidth]{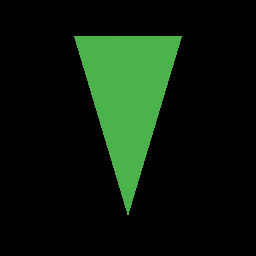}\\[-0.9mm]
    {\footnotesize (a)}
    \end{minipage}
    \begin{minipage}[t] {0.09\textwidth}
    \centering
    \includegraphics[width=\linewidth]{figures/optim/vc3target.png}\\[-0.9mm]
    {\footnotesize (b)}
    \end{minipage}
    \begin{minipage}[t] {0.09\textwidth}
    \centering
    \includegraphics[width=\linewidth]{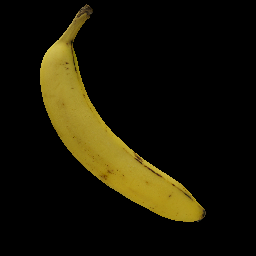}\\[-0.9mm]
    {\footnotesize (c)}
    \end{minipage}
    \begin{minipage}[t] {0.09\textwidth}
    \centering
    \includegraphics[width=\linewidth]{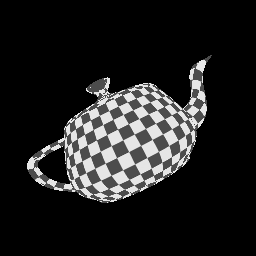}\\[-0.9mm]
    {\footnotesize (d)}
    \end{minipage}
    \begin{minipage}[t] {0.09\textwidth}
    \centering
    \includegraphics[width=\linewidth]{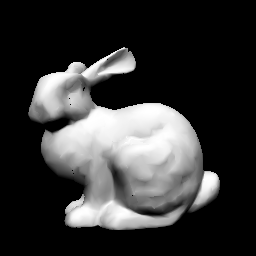}\\[-0.9mm]
    {\footnotesize (e)}
    \end{minipage}
    \begin{minipage}[t] {0.09\textwidth}
    \centering
    \includegraphics[width=\linewidth]{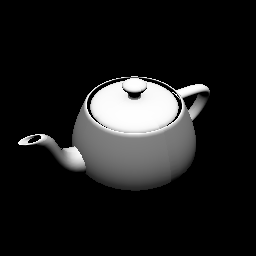}\\[-0.9mm]
    {\footnotesize (f)}
    \end{minipage}
    \begin{minipage}[t] {0.09\textwidth}
    \centering
    \includegraphics[width=\linewidth]{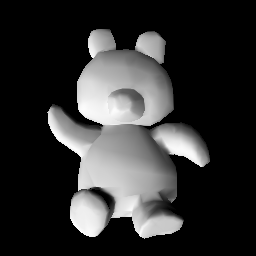}\\[-0.9mm]
    {\footnotesize (g)}
    \end{minipage}
    \begin{minipage}[t] {0.09\textwidth}
    \centering
    \includegraphics[width=\linewidth]{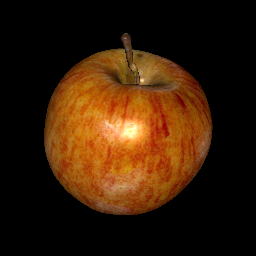}\\[-0.9mm]
    {\footnotesize (h)}
    \end{minipage}
 %       \begin{minipage}[t] {0.09\textwidth}
 %   \centering
 %   \includegraphics[width=\linewidth]{figures/optim2/shininessopt.png}\\[-0.9mm]
 %   {\footnotesize (i)}
 %   \end{minipage}
    \end{minipage}
   % \vspace{-6pt}
   \hspace{-14mm}
   \begin{minipage}{0.35\linewidth}
    \caption{\scriptsize We perform a sanity check for our {\model } by running optimization over several mesh attributes. %\emph{Top row}: initial guesses. \emph{Middle row}: target images. Bottom row, optimization results. 
    We optimize w.r.t. different attributes in different rendering models: (a,b) vertex position and color in the vertex color rendering model, (c,d) texture and texture coordinates in the texture rendering model, (e,f) vertex and camera position in Lambertian model, (g) lighting in the Spherical Harmonic model, (h) material in the Phong model.}
    \label{fig:opt}
    \end{minipage}
    \vspace{-4mm}
\end{figure}

\vspace{-3mm}
\paragraph{Optimization Results.} The design of our differentiable renderer allows for optimization over all defined vertex attributes and a variety of rendering models, which we perform a sanity check for in Fig.~\ref{fig:opt}. % the results of optimizing simultaneously over shape and vertex color in the basic vertex color rendering model, over texture in the texture rendering model, over camera position in the Lambertian model, over lighting in the Spherical Harmonic mode, and over material in the Phong model. Beginning with the initial guess shown in the first row, we optimize over the enumerated parameters and train exclusively through the 2D supervision
Here, we optimize the L-1 loss between the target images (second row) and predicted rendered images. Note that~\cite{li2018differentiable} should be strictly better than us since it supports ray tracing. However, among rasterization-based renderers we are the first to support optimization of all vertex attributes.
%The results of these optimization experiments are shown in the final row, where we demonstrate that we can easily optimize the required mesh properties to perfectly match the provided target. 

\vspace{-7pt}
\section{Applications of {\model }}
\vspace{-6pt}
 We demonstrate the effectiveness of our framework through three challenging ML applications. %For convenience, we denote single view image as $I$, 3D mesh as $M$, 2D texture or vertex color as $C$, and lighting parameters as $L$. We further denote our render as $R(M, C, L) \rightarrow (S, \tilde{I})$, where $S$ is 2D silhouette and $\tilde{I}$ is the colored image.

\vspace{-7pt}
\subsection{Predicting 3D Objects from Single Images}
\label{Single_Img_3D_rec}
\vspace{-6pt}
\paragraph{Geometry and color:} We first apply our approach to the task of predicting a 3D mesh from a single image using only 2D supervision. Taking as input a single RGBA image, with RGB values $I$ and alpha values $S$, a Convolutional Neural Network $F$, parameterized by learnable weights $\theta$, predicts the position and color value for each vertex in a mesh with a specified topology (sphere in our case). We then use a renderer $R$ (specified by shader functions $\Theta$) to render the mesh predicted by $F(I,S;\theta)$ to a 2D silhouette $\tilde{S}$ and the colored image $\tilde{I}$. This prediction pipeline and the architecture details for our mesh prediction network $F$ are provided in Appendix.

When training this system we separate our losses with respect to the silhouette prediction, $\tilde{S}$, and the color prediction, $\tilde{I}$. We use an Intersection-Over-Union (IOU) loss for the silhouette prediction\footnote{We denote  $\mathop{\mathbb{E_{I}}} \triangleq  \mathop{\mathbb{E}_{I\sim p_{data}(I)}}$}: 
\begin{equation}
L_{IOU}(\theta) = \mathbb{E}_{\mathbb{I}} \left[1 - \frac{||S\odot \tilde{S}||_1}{||S + \tilde{S} - S\odot \tilde{S}||_1}\right],
\end{equation}
where $\odot$ denotes element-wise product. Note that $\{\tilde{I},\tilde{S}\}=R(F(I,S;\theta))$ depend on the network's parameters $\theta$ via our {\model}. We further use an L-1 loss for the colored image: 
\begin{equation}
L_{col}(\theta) = \mathbb{E}_{\mathbb{I}} \left[|| I - \tilde{I}||_1\right].
\end{equation}
When rendering our predicted mesh, we not only use the ground truth camera positions and compare against the original image, but also render from a random second view and compare against the ground truth renderings from this new view~\cite{NMR,liu2019soft}. This multi-view loss ensures that the network does not only concentrate on the mesh properties in the known perspective.
We also regularize the mesh prediction with a smoothness loss~\cite{NMR, liu2019soft}, $L_{sm}$, and a Laplacian loss~\cite{liu2019soft,smith19a, wang2018pixel2mesh}, $L_{lap}$, which penalize the difference in normals for neighboring faces and the change in relative positions of neighboring vertices, respectively. The full explanation of these regularizers is provided in Appendix.
The final loss function is then a weighted sum of these four losses:
\begin{eqnarray}
L_{1} = L_{IOU} + \lambda_{col} L_{col} + \lambda_{sm} L_{sm} + \lambda_{lap} L_{lap}.
\vspace{-5pt}
\end{eqnarray}

%\vspace{-5pt}
%\subsection{Texture and Light Prediction}
%\vspace{-5pt}
% in this section, we show we can use our diff-render model to predict texture and light.
% first we talk about how to do it
% the we show the comparison with nm3r
% finally we show with GAN, we can make it more realistic, synthesize multiview, relighting.

% model, texture maping, lighting

 %%%%%%%%%%%%%%%%%%%%%%%%%%%%%%
 % network
 %%%%%%%%%%%%%%%%%%%%%%%%%%%%%%%%%%%%

\begin{figure}[t]
% \vspace{-3mm}
    % \centering
      \begin{minipage}{0.74\linewidth}
   \includegraphics[width=0.9\linewidth]{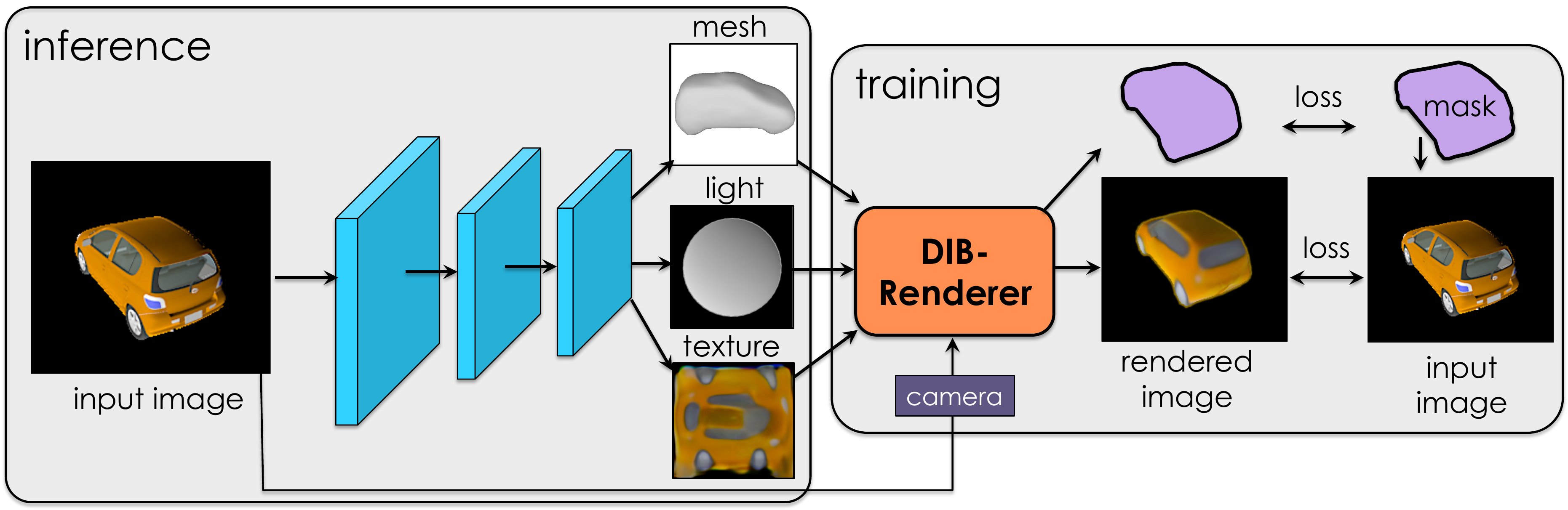}
    \end{minipage}
   % \vspace{-6pt}
   \hspace{-10mm}
   \begin{minipage}{0.322\linewidth}
    \caption{\scriptsize Full architecture of our approach. Given an input image, we predict geometry, texture and lighting. During training we render the prediction with a known camera. We use 2D image loss between input image and rendered prediction to to train our prediction networks. Note that the prediction can vary in different rendering models, e.g.  texture can be vertex color or a texture map while the lighting can be Lambertian, Phong or Spherical Harmonics.}
    \label{fig:net}
    \end{minipage}
    % \vspace{-4mm}
\end{figure}

\paragraph{Geometry, Texture, and Light.} We next apply our method to an extension of the previous task, where a texture map is predicted instead of vertex colors, and lighting parameters are regressed to produce higher quality mesh predictions. %Note that existing rasterization-based renderers are not able to perform this task. \wz{we are comparing with nm3r. How about never tried this task?} 
Our neural network $F$ is modified to predict vertex positions, a texture map, and various lighting information, depending on the lighting model used. Our full learning pipeline is shown in Fig.~\ref{fig:net} and more details are included in the Appendix. We apply the same losses as in the previous section.   %Given an image $I$ and a 2D silhouette $S$, we aim to learn a network $F(I,S;\theta)$ \rightarrow (M, C, L)$, which is trained by rendering $(M, C, L)$ to the corresponding 2D silhouette and colored image $(S, \tilde{I})$ using our {\model } and applying reconstruction loss as provided in the Section.~\ref{Single_Img_3D_rec}. 
 
% While the only supervision is L1 loss and IOU loss\wz{refer to the loss equation in single image reconstruction part} between rendered images and input images, we learn meshes, texture maps and lighting directions in an unsupervised way.
To increase the photo-realism of our predictions, we also leverage an adversarial framework~\cite{goodfellow2014generative, mirza2014conditional}. This is accomplished by training a discriminator network,  $D(\phi)$,  to differentiate between real images, $I$, and rendered mesh predictions, $\tilde{I}$, while our prediction network, $F$, simultaneously learns to make these predictions. 
%on top of the rendered images $\tilde{I}$ and the gradient signal in the loss function back-propagate to the predicted texture map and lighting. 
We adopt the W-GAN \cite{arjovsky2017wasserstein} loss formulation with gradient penalty as in \cite{gulrajani2017improved}: 
\begin{equation}
L_{adv}(\theta, \phi)= \mathop{\mathbb{E_{I}}}\left[D(I;\phi)  - D(\tilde{I};\phi)\right],\quad
L_{gp}(\phi) =\mathop{\mathbb{E_{\tilde{I}}}}\left[(||\nabla_{\tilde{I}} D(\tilde{I};\phi)||_2  - 1)^2\right].
\end{equation}
Similar to~\cite{isola2017image, wang2018high, yao20183d}, we additionally use a perceptual loss and discriminator feature matching loss to make training more stable: 
\begin{equation} 
L_{percep}(\phi) =  \mathop{\mathbb{E_{I}}}\left[\sum_{i=1}^{M_{V}} \frac{1}{N_i^V} || V^i(I) - V^i( \tilde{I}) ||_1  + \sum_{i=1}^{M_{D}} \frac{1}{N_i^D} || D^i(I;\phi) - D^i( \tilde{I};\phi) ||_1 \right], 
\end{equation}
where $V^i$ denotes the $i$-th layer of a pre-trained VGG network, $V$, with $N_i^V$ elements, $D^i$ denotes the $i$-th layer in the discriminator $D$ with $N_i^D$ elements, and the numbers of layers in network $V$ and $D$ are $M_{V}$ and $M_{D}$, respectively. Our full objective function for this task is then: 
\begin{equation}
\theta^*, \phi^* = \argmin \limits_{\theta}\bigg(\argmax\limits_{\phi}\left( \lambda_{adv} L_{adv} - \lambda_{gp}L_{gp}\right) +  \lambda_{per} L_{percep}  +  L_{1}\bigg).
\end{equation}

% \begin{eqnarray}
% L_{adv}(\theta, \phi)&=& \mathop{\mathbb{E_{I}}}\left[D(I;\phi)  - D(\tilde{I};\phi)\right]  \\
% L_{gp}(\phi) &=&\mathop{\mathbb{E_{\tilde{I}}}}\left[(||\nabla_{\tilde{I}} D(\tilde{I};\phi)||_2  - 1)^2\right]\\
% L_{percep}(\phi) &=&  \mathop{\mathbb{E_{I}}}\left[\sum_{i=1}^{M_{V}} \frac{1}{N_i^V} || V^i(I) - V^i( \tilde{I}) ||_1  + \sum_{i=1}^{M_{D}} \frac{1}{N_i^D} || D^i(I;\phi) - D^i( \tilde{I};\phi) ||_1 \right]\\
% \theta^*, \phi^*& =& \argmin \limits_{\theta}\bigg(\argmax\limits_{\phi}\left( \lambda_{adv} L_{adv} - \lambda_{pen}L_{penalty}\right) +  \lambda_{per} L_{percep}  +  L_{1}\bigg)
% \end{eqnarray}

% \textbf{Texture Maps and Lighting Directions}
% % To render with texture maps, we need to decide the corresponding 2D uv coordinates.  We adopt a UNet\cite{} to map the input image to the texture maps. Besides, we predict lighting directions as well, whe

% We adopt a UNet\cite{} to map a 256x256 input image to a 256x256 texture map. Since we deform meshes from a sphere template, similar to \cite{cmrKanazawa18}, we use 2D spherical coordinates as the UV coordinates. We also predict the corresponding lighting direction, where we apply a 6-layer ResNet\cite{} to regress the XYZ directions from the UNet features. To learn the same distribution, our model are trained with shapenet dataset rendered with Lambertian model as well.

% We show the qualitative and quantitative comparison in Fig .\ref{}.

% \paragraph{Objective Function}

\vspace{-10pt}
\subsection{3D GAN of Textured Shapes via 2D supervision}
\vspace{-5pt}
In our second application, we further demonstrate the power of our method by training a Generative Adversarial Network (GAN)~\cite{goodfellow2014generative} to produce 3D textured shapes using only 2D supervision. We train a network $F_{GAN}$ to predict vertex positions and a texture map, and exploit a discriminator $D(\phi)_{I}$ to differentiate between real images, and rendered predictions. The network $F_{GAN}$ is modified so as to take normally distributed noise as input, in place of an image. 

While empirically the above GAN is able to recover accurate shapes, it fails to produce meaningful textures. We suspect that disentangling both shape and texture by an image-based discriminator is a hard learning task. To facilitate texture modeling, we train a second discriminator $D(\sigma)_t$, which operates over texture map predictions. However, as our dataset does not contain true texture maps which can be mapped onto a deformed sphere, for ground truth textures we instead use the textures produced from our network trained to predict texture and lighting from images (Sec~\ref{Single_Img_3D_rec}). To produce these texture maps, we pass every image in our training set through our texture and lighting prediction network, and extract the predicted texture. Our second discriminator then learns to differentiate between textures generated by $F_{GAN}$, and the extracted learned textures. We train $F_{GAN}$ via W-GAN with gradient penalty~\cite{gulrajani2017improved}, and use both discriminators to provide a learning signal.  

%not sure of the difference between evaluation adn application, can we just call it all experiments?

\vspace{-6pt}
\section{Experiments}
\vspace{-10pt}
%\vspace{-5pt}
\begin{table}[t!]
\vspace{-3mm}
	\begin{center}
	{
	\caption{Results on single image 3D object prediction reported with 3D IOU (\%) / F-score (\%).}
% 	\vspace{-7pt}
		\label{tbl:3diou}
		\begin{adjustbox}{width=1.01\linewidth}
		\addtolength{\tabcolsep}{-3.9pt}
		\hspace{-4mm}\begin{tabular}{c|c|c|c|c|c|c|c|c|c|c|c|c|c|c}
			\toprule
		    Category & Airplane & Bench & Dresser & Car & Chair &Display &Lamp & Speaker & Rifle & Sofa & Table & Phone & Vessel & Mean\\
		    \toprule
		    N3MR~\cite{NMR} &\textbf{ 58.5/80.6}&45.7/55.3&74.1/46.3&71.3/53.3&41.4/39.1&55.5/43.8&36.7/\textbf{46.4}&67.4/35.0&55.7/\textbf{83.6} & 60.2/39.2 & 39.1/46.9 & \textbf{76.2/74.2} & 59.4/\textbf{66.9} & 57.0/54.7\\
			SoftR.~\cite{liu2019soft} &58.4/71.9&44.9/49.9&73.6/41.5&77.1/51.1&49.7/40.8&54.7/41.7&39.1/39.1&68.4/29.8& \textbf{62.0}/82.8 & 63.6/39.3 & 45.3/37.1 & 75.5/68.6 & 58.9/55.4 & 59.3/49.9\\
			%Our (sil) &0.5643&0.4953&0.7652&0.7859&0.5232&0.5911&0.4035&\\
			Ours &57.0/75.7&\textbf{49.8/55.6}&\textbf{76.3/52.2}&\textbf{78.8/53.6}&\textbf{52.7/44.7}&\textbf{58.8/46.4}&\textbf{40.3}/45.9&\textbf{72.6/38.8}&56.1/82.0 & \textbf{67.7/43.1} & \textbf{50.8/51.5} & 74.3/73.3 &  \textbf{60.9}/63.2 & \textbf{61.2/55.8}\\
			\bottomrule
		\end{tabular}
		\end{adjustbox}
		}
	\end{center}
	\vspace{-6mm}
\end{table}

\vspace{-1mm}
\paragraph{Dataset: }
%\vspace{-5pt}
As in~\cite{NMR,liu2019soft,wang2018pixel2mesh}, our dataset comprises 13 object categories from the ShapeNet dataset~\cite{ShapeNet}. We use the same split of objects into our training and test set as~\cite{wang2018pixel2mesh}. We render each model from 24 different views to create our dataset of RGB images used for 2D supervision. To demonstrate the multiple rendering models which {\model } supports, we render each image with 4 different rendering models: 1) basic model without lighting effects, 2) with Lambertian reflectance, 3) with Phong shading, and 4) with  Spherical Harmonics. Further details are provided in the Appendix.
%A comprehensive description of dataset preparation for each experiment is provided in the Appendix. %\jun{shall we mention we will release the dataset and code in some place?}

\vspace{-6pt}
\subsection{Predicting 3D Objects from Single Images: Geometry and Color}
\vspace{-5pt}
\paragraph{Experimental settings:} In our experiments,  we set $\lambda_{col}=1, \lambda_{sm}=0.001,$ and $\lambda_{lap}=0.01$. We train one network on all 13 categories. The network is optimized using the Adam optimizer~\cite{kingma2014adam}, with $\alpha=0.0001, \beta_1 = 0.9$, and $\beta_2=0.999$. The batch size is 64, and the dimension of input image is $64\times64$.  We compare our method with the two most related differentiable renderers, N3MR~\cite{NMR} and SoftRas-Mesh~\cite{liu2019soft}, using the same network configurations, training data split and hyperparameters. For quantitative comparison, we first voxelize the predicted mesh into $32^3$ volume using a standard voxelization tool \texttt{binvox}\footnote{http://www.patrickmin.com/binvox/} provided by ShapeNet~\cite{ShapeNet} and then evaluate using 3D IOU, a standard metric in 3D reconstruction. We additionally measure the F-score following~\cite{tatarchenko2019single} between the predicted mesh and ground truth mesh. The tolerance for F-score is set to 0.02.

\begin{figure}
    \centering
    \begin{minipage}[t] {0.115\textwidth}
    \centering
    \includegraphics[width=\linewidth, height=0.75\linewidth, trim=100 100 100 100, clip]{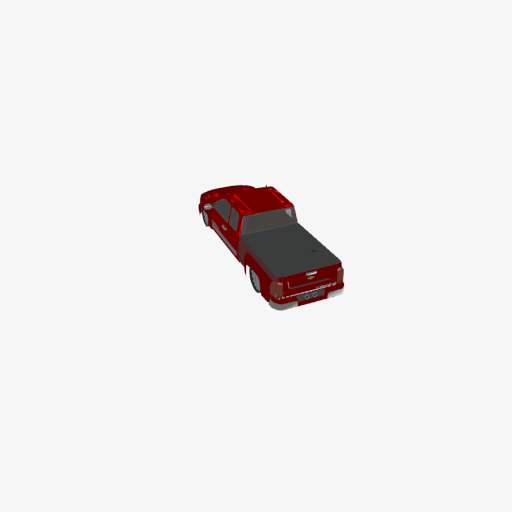}
    \end{minipage}
    \begin{minipage}[t] {0.115\textwidth}
    \centering
    \includegraphics[width=\linewidth, height=0.75\linewidth, trim=100 100 100 100, clip]{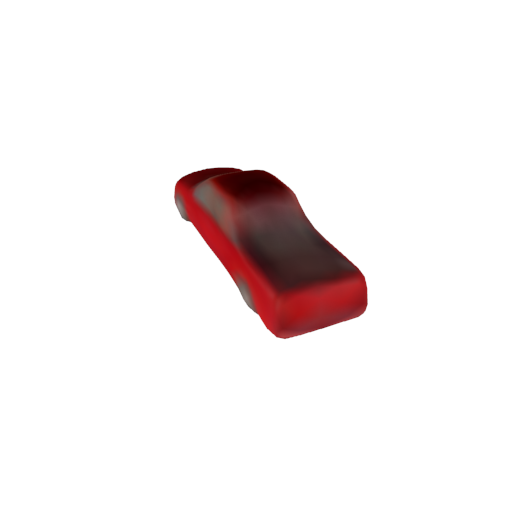}
    \end{minipage}
    \begin{minipage}[t] {0.115\textwidth}
    \centering
    \includegraphics[width=\linewidth, height=0.75\linewidth, trim=100 100 100 100, clip]{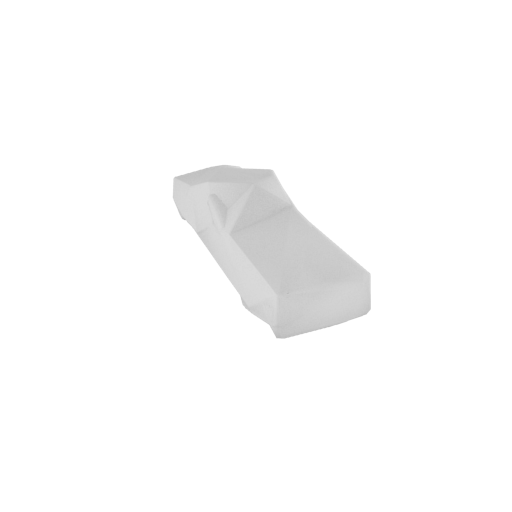}
    \end{minipage}
    \begin{minipage}[t] {0.115\textwidth}
    \centering
    \includegraphics[width=\linewidth,height=0.75\linewidth, trim=100 100 100 100, clip ]{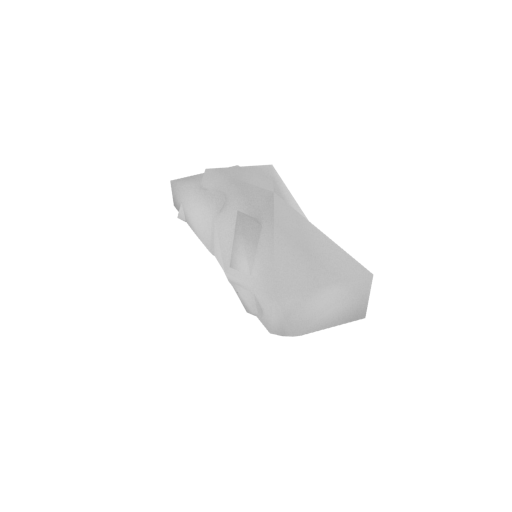}
    \end{minipage}
    \begin{minipage}[t] {0.115\textwidth}
    \centering
    \includegraphics[width=\linewidth, height=0.75\linewidth, ]{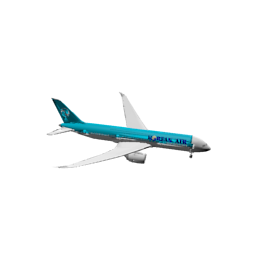}
    \end{minipage}
    \begin{minipage}[t] {0.115\textwidth}
    \centering
    \includegraphics[width=\linewidth, height=0.75\linewidth, trim=100 100 100 100, clip]{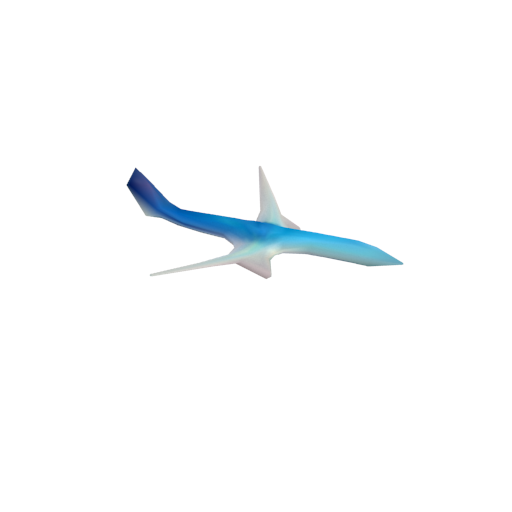}
    \end{minipage}
    \begin{minipage}[t] {0.115\textwidth}
    \centering
    \includegraphics[width=\linewidth, height=0.75\linewidth, trim=100 100 100 100, clip]{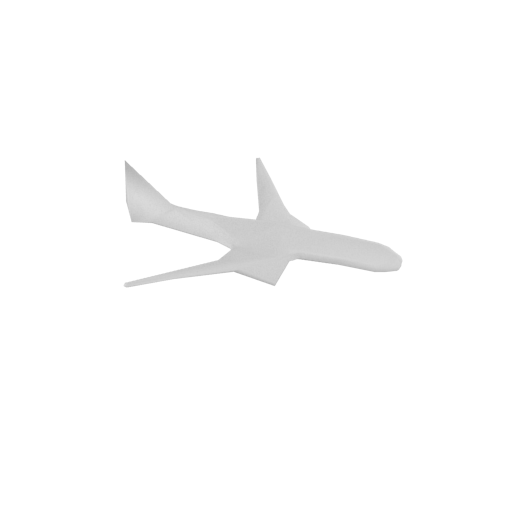}
    \end{minipage}
    \begin{minipage}[t] {0.115\textwidth}
    \centering
    \includegraphics[width=\linewidth, height=0.75\linewidth, trim=100 100 100 100, clip]{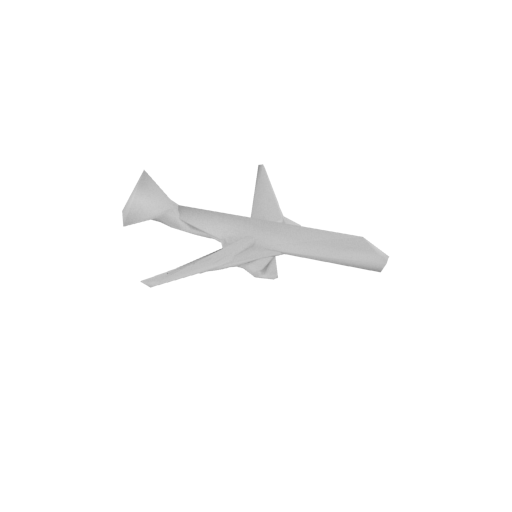}
    \end{minipage}
    \begin{minipage}[t] {0.115\textwidth}
    \centering
    \includegraphics[width=\linewidth, height=0.75\linewidth, trim=100 100 100 100, clip ]{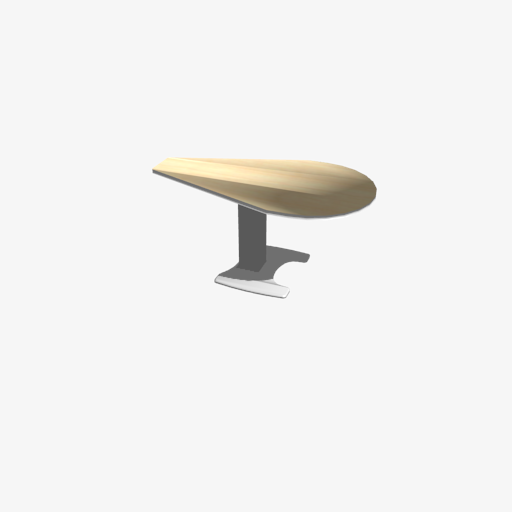}
    \end{minipage}
    \begin{minipage}[t] {0.115\textwidth}
    \centering
    \includegraphics[width=\linewidth, height=0.75\linewidth, trim=100 100 100 100, clip]{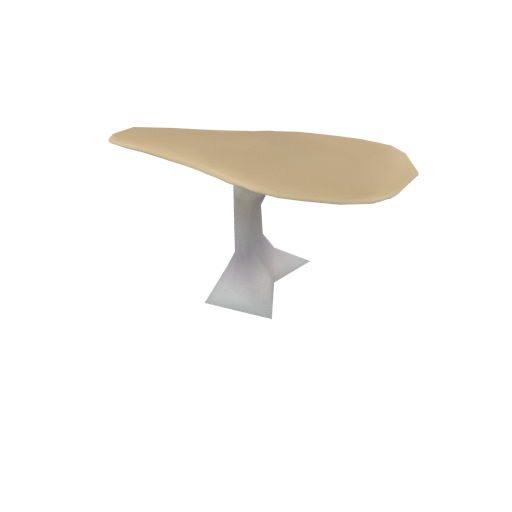}
    \end{minipage}
    \begin{minipage}[t] {0.115\textwidth}
    \centering
    \includegraphics[width=\linewidth, height=0.75\linewidth, trim=100 100 100 100, clip]{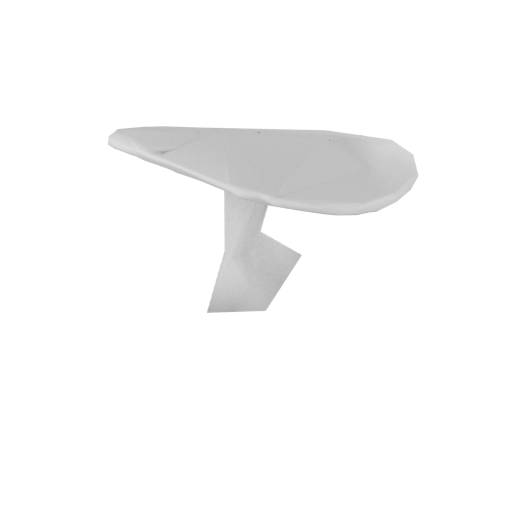}
    \end{minipage}
    \begin{minipage}[t] {0.115\textwidth}
    \centering
    \includegraphics[width=\linewidth, height=0.75\linewidth, trim=100 100 100 100, clip]{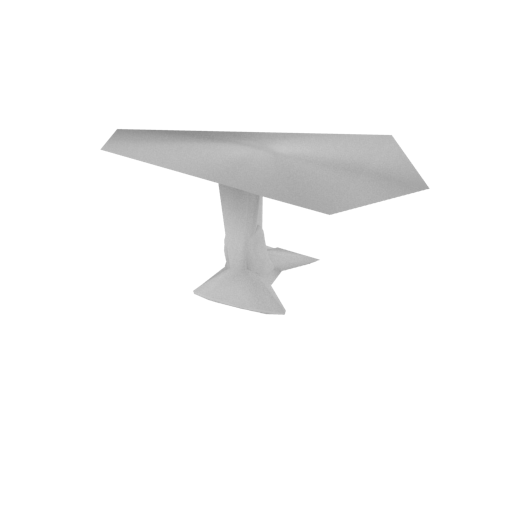}
    \end{minipage}
    \begin{minipage}[t] {0.115\textwidth}
    \centering
    \includegraphics[width=\linewidth, height=0.75\linewidth,  trim=100 100 100 100, clip ]{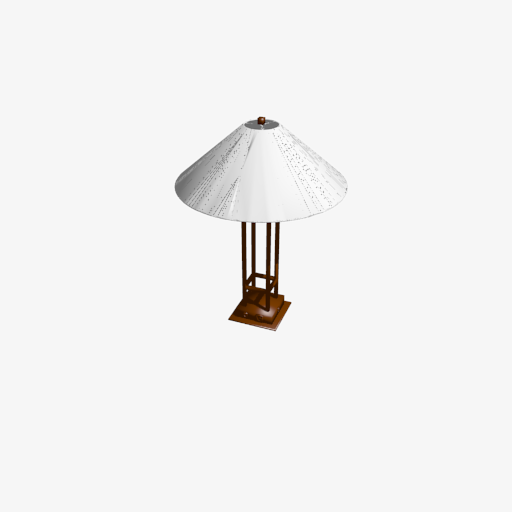}
    \end{minipage}
    \begin{minipage}[t] {0.115\textwidth}
    \centering
    \includegraphics[width=\linewidth, height=0.75\linewidth, trim=100 10 100 10, clip]{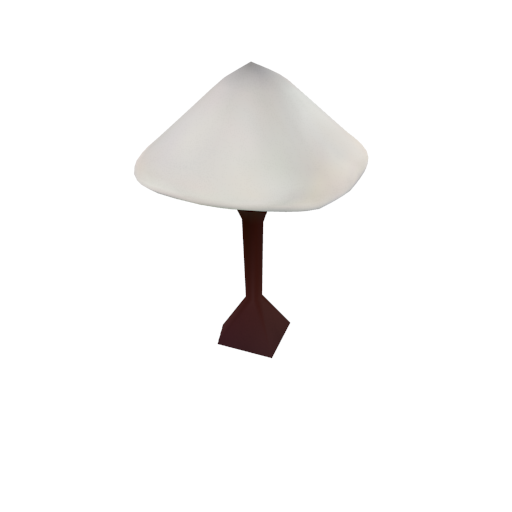}
    \end{minipage}
    \begin{minipage}[t] {0.115\textwidth}
    \centering
    \includegraphics[width=\linewidth, height=0.75\linewidth, trim=100 10 100 10, clip]{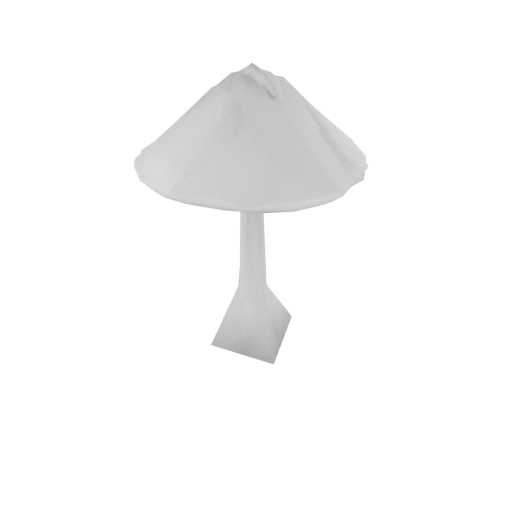}
    \end{minipage}
    \begin{minipage}[t] {0.115\textwidth}
    \centering
    \includegraphics[width=\linewidth, height=0.75\linewidth, trim=100 10 100 10, clip]{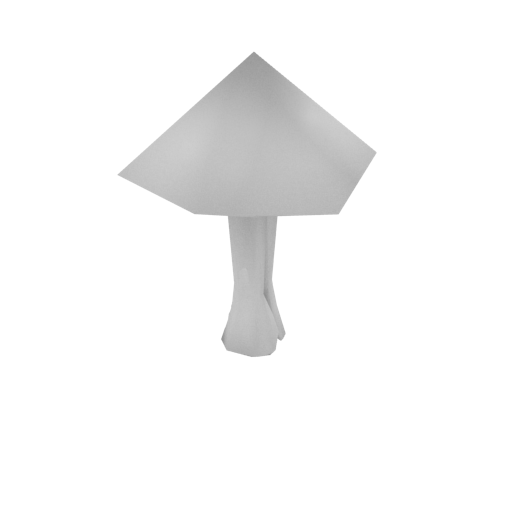}
    \end{minipage}
    \vspace{-15pt}
    \caption{\footnotesize Qualitative results on single image 3D object prediction. First and fifth column is the ground-truth image, the second and sixth columns are the prediction from our model, the third and seventh column are results from SoftRas-Mesh~\cite{liu2019soft}, the rest two columns are results from N3MR~\cite{NMR}.} 
    \vspace{-0.0em}
    \label{fig:3d-reconstruction}
\end{figure}

\vspace{-5pt}
\paragraph{Results:} Table~\ref{tbl:3diou} provides an evaluation. Our {\model} significantly outperforms other methods, in almost all categories on both metrics. We surpass SoftRas-Mesh/N3MR with 1.92/4.23 points and 5.98/1.23 points in terms of 3D IOU and F-score, respectively. As the only difference in this experiment is the renderer, the quantitative results demonstrate the superior performance of our method. Qualitative examples are shown in Fig~\ref{fig:3d-reconstruction}. Our {\model} faithfully reconstructs both the fine-detailed color and the geometry of the 3D shape, compared to SoftRas-Mesh and N3MR. %Additional results are in Appendix.

\vspace{-8pt}
\subsection{Predicting 3D Objects from Single Images: Geometry, Texture and Light}
\vspace{-7pt}

\paragraph{Experimental settings:}

% first, compare with nm3r in Lambertian model
We adopt a UNet~\cite{ronneberger2015u} architecture to predict texture maps. As we deform a mesh from a sphere template, similar to \cite{kanazawa2018learning}, we use 2D spherical coordinates as the UV coordinates. The dimension of the input image and predicted texture is $256\times256$.  We use a 6-layer ResNet~\cite{he2016deep} architecture to regress the XYZ directions of light at each vertex from the features at the bottleneck layer of UNet network. We only use the Lambertian model and qualitatively compare with N3MR~\cite{NMR} by measuring the reconstruction error on rendered images under identical settings. Note that SoftRas-Mesh~\cite{liu2019soft} does not support texture and lighting and so no comparison can be made. In the following sections, we only perform experiments on the car class, which has more diverse texture.

% We also demonstrate our render with spherical harmonic model. To further improve the quality of our prediction, we try to add an Adversial Loss(see Sec.\ref{} ) to learn more detials. We show with such a loss, we can predict much more faithful results.

\begin{table*}
\vspace{-3.5mm}
\begin{minipage}{0.45\linewidth}
\vspace{-3mm}
    \begin{center}
%   \vspace{-7pt}
%       \label{tbl:texture}
        {\small
        \addtolength{\tabcolsep}{-2.5pt}
        \begin{tabular}{cccc}
            \toprule
            Models &Texture&Lighting & Text. + Light \\
            \toprule
            N3MR~\cite{NMR} & 0.03640 & 23.5585 &0.02208\\
            Ours &\textbf{0.02179}&\textbf{9.7096}&\textbf{0.01362}\\
            \bottomrule
        \end{tabular}
        }
    \end{center}
    \end{minipage}
    \hspace{3mm}
    \begin{minipage}{0.52\linewidth}
        
    %   \vspace{-3mm}
    \caption{\footnotesize Results for texture and light prediction. Texture/Texture+Light shows L-1 loss on the rendered image for texture/texture+lighting. Lighting shows the angle between predicted lighting and GT lighting. Lower is better.}
    \label{tbl:texture}
        \end{minipage}
        \vspace{-1mm}
\end{table*}

\begin{figure*}[t!]
\vspace{-4.5mm}
\includegraphics[height=1.57cm,trim=0 0 0 0, clip]{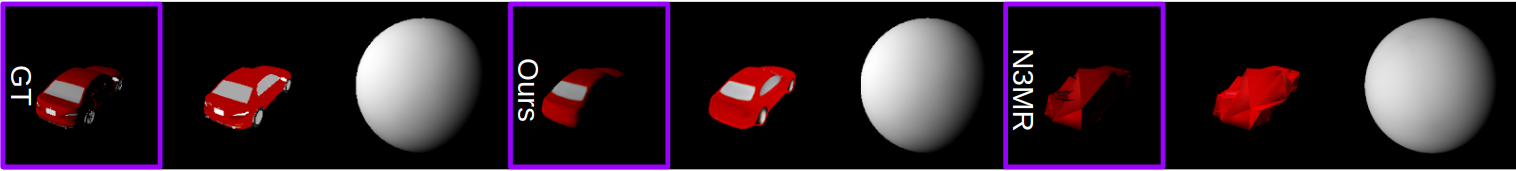}
\vspace{-5.5mm}
\caption{\footnotesize{Qualitative examples for 3D shape, texture and light prediction. Col. 1-3: 1) GT rendered image with texture+light, 2) texture only rendered image, 3) light map. Col 4-6: our predictions. Col: 7-9: N3MR~\cite{NMR}}}
\label{fig:comp_nm3r}
\vspace{-4mm}
\end{figure*}

\vspace{-6pt}
\paragraph{Results:} We provide results in Table~\ref{tbl:texture} and Fig~\ref{fig:comp_nm3r}. As ShapeNet does not provide ground-truth UV texture, we compute the L-1 difference on the rendered image using the predicted texture/texture+lighting and the GT image. Compared to N3MR~\cite{NMR}, we achieve significantly better results both quantitatively and qualitatively. We obtain about 40\% lower L-1  difference on texture and 60\% smaller angle difference on lighting direction than N3MR. We also obtain significantly better visual results, in terms of the shape, texture and lighting, as shown in Fig~\ref{fig:comp_nm3r} and the Appendix.

\vspace{-5pt}
\subsection{Texture and Lighting Results With Adversarial Loss}
\vspace{-5pt}
We now evaluate the effect of adding the adverserial loss to the previous experiment.  %when training a network to predict geometry, texture and lighting. 
We demonstrate our DIB-Render with Phong and Spherical Harmonic lighting models. For Phong model we keep diffuse and specular reflectance constant as 1 and 0.4 respectively and predict lighting direction together with shininess constant $\alpha$ while for Spherical Harmonic model we predict 9 coefficients.

\begin{figure*}[t!]
% \vspace{-3mm}
\addtolength{\tabcolsep}{-3.6pt}
\includegraphics[height=2.43cm,trim=0.35cm 5cm 0 5cm,clip]{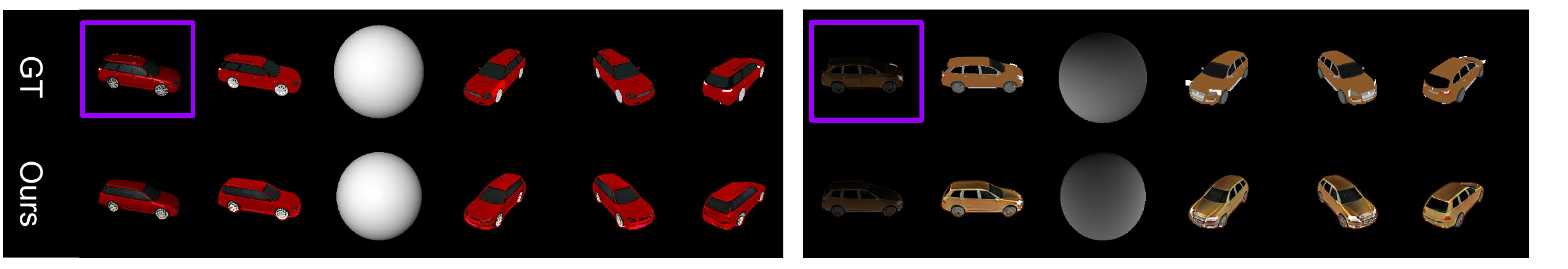}
\vspace{-6.5mm}
\caption{\footnotesize{Qualitative examples for 3D shape, texture and light prediction, when exploiting {\bf adverserial loss}}.  {\bfseries Purple rectangle}: Input image.
{\bfseries Left Example}: Phong Lighting. {\bfseries Right Example}: Spherical Harmonics. 
{\bfseries First col}: Texture and Light. {\bfseries Second col}: Texture. {\bfseries Third col}: Light. {\bfseries Forth to Sixth col}: Texture; different views.}
\label{fig:quantitive_results1}
\vspace{-3.5mm}
\end{figure*}

\vspace{-4mm}
\paragraph{Experimental settings:} We first train the model without adversarial loss for 50000 iterations then  fine-tune it with adversarial loss for extra 15000 steps. The detailed network architecture is provided in the Appendix. We set $\lambda_{adv}=0.5, \lambda_{gp}=0.5,$ and $\lambda_{per}=1$. We fix the learning rate for the discriminator to $1e^{-5}$ and optimize using Adam~\cite{kingma2014adam}, with $\alpha=0.0001, \beta_1 = 0.5$, and $\beta_2=0.999$.
\vspace{-5pt}
\vspace{-3mm}
\paragraph{Qualitative Results and Separation Study:}
We show qualitative results in Fig.~\ref{fig:quantitive_results1}. Notice that the network disentangles texture and light quite well, and produces accurate 3D shape. Furthermore, the adverserial loss helps in making the predicted texture look more crisp compared to Fig~\ref{fig:comp_nm3r}. %Our model also learn to disentangle texture and lighting in an unsupervised manner, where the supervision only exists in input image and rendered images. To better understand the disentanglement of all the predictions, we do the ablation on images with different settings, where we change one setting while keeping others fixed. 

\begin{figure*}[t!]
\vspace{-1mm}
\addtolength{\tabcolsep}{-3.6pt}
\hspace{-1mm}\includegraphics[width=\textwidth,trim=1cm 5.2cm 1cm 4.6cm,clip]{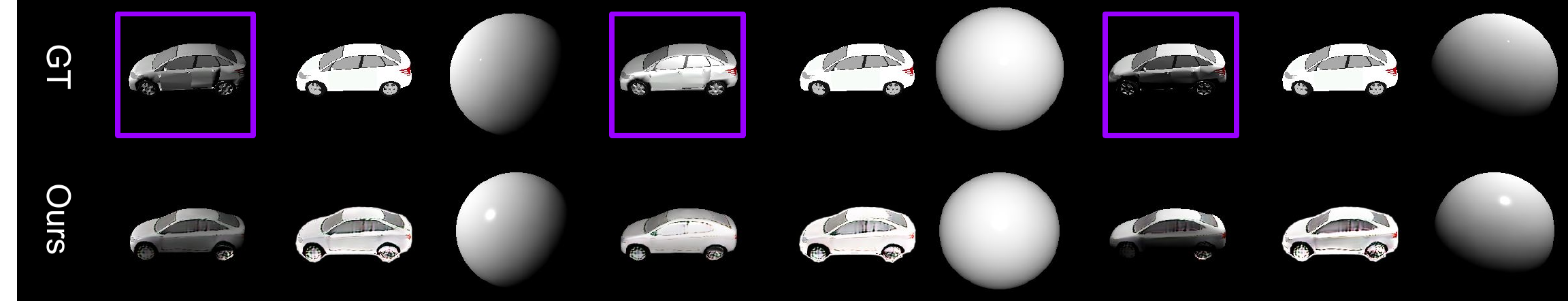}\\
\vspace{-6mm}
\caption{\footnotesize{Light \& Texture Separation Study}. {\bfseries Purple rectangle}: Input image, which are rendered with the same car model but different lighting directions. Each three columns visualize \bf{Texture + Light, Texture, Light}.}
\label{fig:quantitive_results2}
\includegraphics[width=\textwidth,trim=0.2cm 5.2cm 0.2cm 4.5cm,clip]{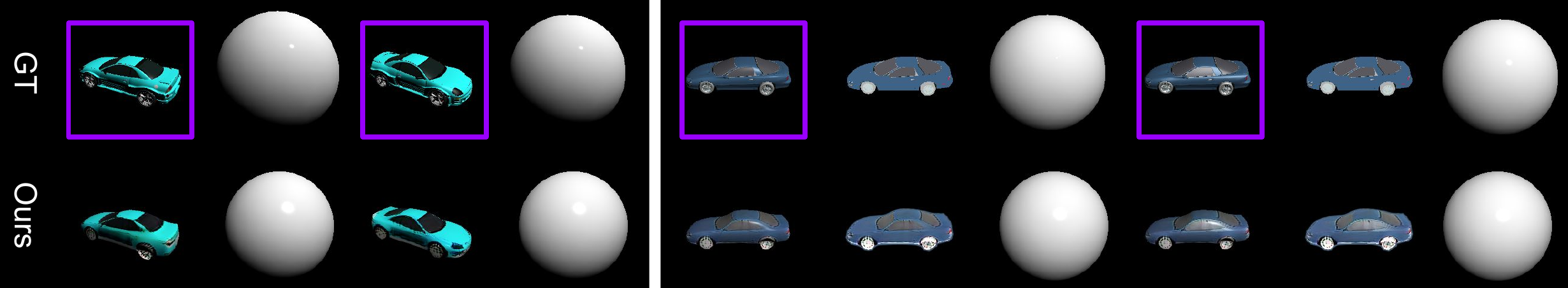}
\vspace{-6mm}
\caption{\footnotesize{Light \& Texture Separation Study}. {\bfseries Purple rect}: Input image. {\bfseries Left:} Input images are with same light and texture but vary views.  {\bfseries Right:}  Input images are with the same texture but with different shininess constants.}
\label{fig:quantitive_results3}
\vspace{-1mm}
\end{figure*}

To further study texture and light disentanglement, we render test input images with the same car model but vary the lighting direction (Fig.~\ref{fig:quantitive_results2}). Our predictions recover the shape, texture map and  lighting directions. 
Further examples are provided in Fig.~\ref{fig:quantitive_results3}. On the left, we fix the lighting and texture but render the car in different camera views, to illustrate consistency of prediction across viewpoints. On the right of the figure, we render images with different shininess constants, but fixed lighting direction, texture and camera view. Here, we find that our model is not able to accurately predict the shininess constant. 
In this case, the texture map erroneously compensates for the shininess effect. This might be because the shininess effect is not significant enough to be learned by a neural network though 2D supervision. We hope to resolve this issue in future work.

\begin{table*}
\vspace{-3.5mm}
\begin{minipage}{0.45\linewidth}
\vspace{-3mm}
	\begin{center}
% 	\vspace{-7pt}
		{\small
		\addtolength{\tabcolsep}{-2.5pt}
		\begin{tabular}{cccc}
			\toprule
		    Models &Texture& 2D IOU & Key Point \\
		    \toprule
		    CMR~\cite{kanazawa2018learning} & 0.043 & 0.262 &0.930\\
			Ours & 0.043 &\textbf{0.243}&\textbf{0.972}\\
			\bottomrule
		\end{tabular}
		}
	\end{center}
	\end{minipage}
	\hspace{3mm}
	\begin{minipage}{0.52\linewidth}
		
	%	\vspace{-3mm}
	\caption{\footnotesize Results on CUB bird dataset~\cite{cub}. Texture and 2D IOU show L-1 loss and 2D IOU loss between predictions and GT, lower is better. Key point evaluates percentage of predicted key points lying in the threshold of 0.1, higher is better.}
	\label{tbl:real}
		\end{minipage}
		\vspace{-1mm}
\end{table*}

\begin{figure*}[t!]
\vspace{-2.5mm}
\includegraphics[height=1.56cm,trim=0 0 0 0, clip]{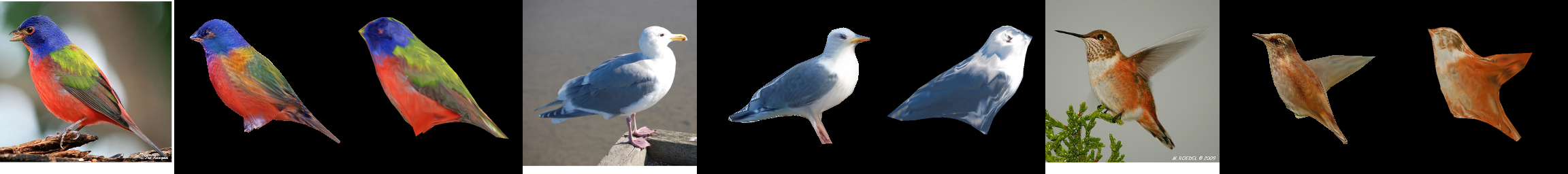}
\includegraphics[height=1.56cm,trim=0 0 0 0, clip]{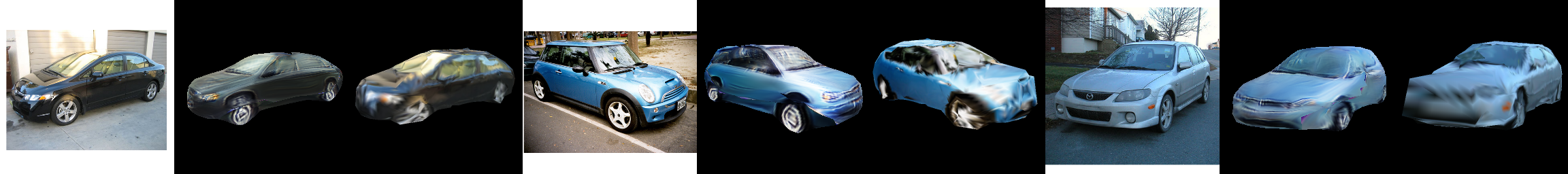}
\begin{center}
\vspace{-5mm}
\begin{small}
\begin{tabular}{p{11.4mm}p{11.4mm}p{11.4mm}p{11.4mm}p{11.4mm}p{11.4mm}p{11.4mm}p{11.4mm}p{11.4mm}}
\hspace{1.5mm} input & \hspace{1.8mm} ours & \hspace{1mm} CMR & \hspace{1.5mm} input & \hspace{1.8mm} ours & \hspace{1mm} CMR & \hspace{1.5mm} input & \hspace{1.8mm} ours & \hspace{1mm} CMR 
\end{tabular}
\end{small}
\end{center}
\vspace{-4.0mm}
\caption{\footnotesize{% Qualitative examples for real images shape and texture prediction on CUBbird dataset~\cite{cub}  and PASCAL3D+ Car dataset~\cite{pascal3d}. We compare our results with CMR~\cite{kanazawa2018learning}.
Qualitative examples on CUBbird dataset~\cite{cub}  and PASCAL3D+ Car dataset~\cite{pascal3d}}}
%Col. (1,4,7): Input image. Col. (2, 5, 8): Our predictions. Col. (3,6,9): CMR~\cite{cmrKanazawa18}.}}
\label{fig:comp_cmr}
\vspace{-4.0mm}
\end{figure*}

\subsection{Real Images}
\vspace{-1.5mm}

\paragraph{Experimental settings:}  We next test our method on real images. Since real images generally do not have multiview captures for the same object, it would be very hard to infer precise light because light and texture would be entangled together. Following CMR~\cite{kanazawa2018learning}, we adopt CUB bird dataset~\cite{cub} and PASCAL3D+ car dataset~\cite{pascal3d}, predicting shape and texture from a single view image.

\vspace{-3mm}
\paragraph{Results:}
We compare our method with CMR~\cite{kanazawa2018learning}. Instead of predicting texture flow, we predict texture map directly. Both two methods use GT cameras estimated from structure from motion. Table \ref{tbl:real} provides quantitative evaluation of predicted texture and shape on CUB bird dataset. Our show better shape and key points predictions than CMR. While the loss of texture predictions are the same, Fig. \ref{fig:comp_cmr} shows qualitative improvements our method provides. Our textures are of higher fidelity and more realistic. This is because we predict a whole image as the texture map while CMR~\cite{kanazawa2018learning} adopts N3MR~\cite{NMR}, which uses face color as the texture and so the restricted face size results in blurriness. For car prediction, both two methods clearly posses poor artifacts. This is because the segmentation in PASCAL3D+ car dataset is estimated from Mask R-CNN~\cite{maskrcnn}, which is not perfect. In addition, the car textures have more details than birds, which make it very hard to learn good shape and texture. Despite these facts, the visual quality our of predictions continues to display a marked improvement. % We leave it for future work.

\begin{figure*}[t!]
\vspace{-0.5mm}
\centering
\includegraphics[width=\textwidth,trim=0 0 490 10,clip]{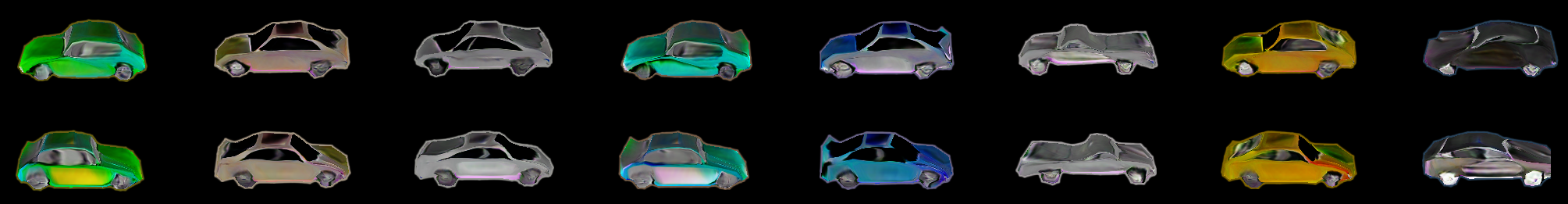}
\vspace{-6mm}
\caption{{\footnotesize Samples from our 3D GAN trained on car images, from 2 viewpoints (shown in each column).}}
\label{GAN_sample}
\vspace{-2mm}
\end{figure*}

\begin{figure*}[t!]
\vspace{-2.5mm}
\includegraphics[width=\textwidth,trim=0 0 490 0,clip]{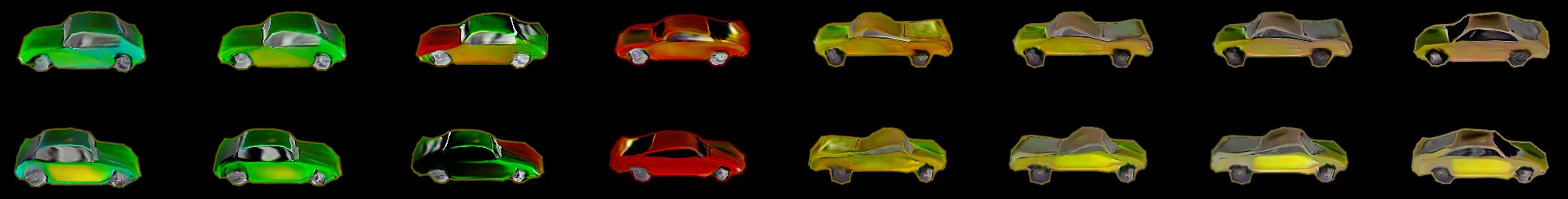}
\vspace{-5.5mm}
\caption{\footnotesize Renderings of objects produced by interpolating between latent codes of our 3D GAN, from 2 views}
\label{GAN_inter}
\vspace{-6mm}
\end{figure*}

\vspace{-3mm}
\subsection{3D GAN of Textured Shapes via 2D Supervision}
\vspace{-1.5mm}
\paragraph{Experimental settings:}  We first train the networks to only predict shape, by only passing gradients back through the silhouette prediction. We then fix the shape prediction and only train to produce textures. We perform this experiment on the car class. Images from 4 primary views are rendered for each predicted mesh in each training iteration, and concatenated together when passed to rendered image discriminator to force generation of objects in a canonical pose.

\vspace{-3mm}
\paragraph{Results:}
We show the results of randomly sampling from our learned distribution of car shapes and textures in Fig~ \ref{GAN_sample}. This figure demonstrates the high quality of of shape and texture generations, in addition to their diversity. We also show the result of rendering meshes produced from interpolation between latent codes in Fig~\ref{GAN_inter}, to demonstrate the robust nature of our learned distribution. %Additional results are provided in Appendix. 

\vspace{-5mm}
\section{Conclusion}
\vspace{-4mm}
In this paper, we proposed a complete rasterization-based differentiable renderer for which gradients can be computed analytically. Our framework, when wrapped around a neural network, learns to predict shape, texture, and light from single images. We further showcase our framework to learn a generator of 3D textured shapes.

\small
\paragraph{Acknowledgement}
\vspace{-4mm}
Wenzheng Chen wants to thank the support of DARPA under the REVEAL program and NSERC under the COHESA Strategic Network.

\bibliography{main}

\begin{thebibliography}{10}

\bibitem{achlioptas2017learning}
Panos Achlioptas, Olga Diamanti, Ioannis Mitliagkas, and Leonidas Guibas.
\newblock Learning representations and generative models for 3d point clouds.
\newblock {\em arXiv preprint arXiv:1707.02392}, 2017.

\bibitem{arjovsky2017wasserstein}
Martin Arjovsky, Soumith Chintala, and L{\'e}on Bottou.
\newblock Wasserstein gan.
\newblock {\em arXiv preprint arXiv:1701.07875}, 2017.

\bibitem{ShapeNet}
Angel~X Chang, Thomas Funkhouser, Leonidas Guibas, Pat Hanrahan, Qixing Huang,
  Zimo Li, Silvio Savarese, Manolis Savva, Shuran Song, Hao Su, et~al.
\newblock Shapenet: An information-rich 3d model repository.
\newblock {\em arXiv preprint arXiv:1512.03012}, 2015.

\bibitem{genova2018unsupervised}
Kyle Genova, Forrester Cole, Aaron Maschinot, Aaron Sarna, Daniel Vlasic, and
  William~T Freeman.
\newblock Unsupervised training for 3d morphable model regression.
\newblock In {\em Proceedings of the IEEE Conference on Computer Vision and
  Pattern Recognition}, pages 8377--8386, 2018.

\bibitem{goodfellow2014generative}
Ian Goodfellow, Jean Pouget-Abadie, Mehdi Mirza, Bing Xu, David Warde-Farley,
  Sherjil Ozair, Aaron Courville, and Yoshua Bengio.
\newblock Generative adversarial nets.
\newblock In {\em Advances in neural information processing systems}, pages
  2672--2680, 2014.

\bibitem{greene1993hierarchical}
Ned Greene, Michael Kass, and Gavin Miller.
\newblock Hierarchical z-buffer visibility.
\newblock In {\em Proceedings of the 20th annual conference on Computer
  graphics and interactive techniques}, pages 231--238. ACM, 1993.

\bibitem{gulrajani2017improved}
Ishaan Gulrajani, Faruk Ahmed, Martin Arjovsky, Vincent Dumoulin, and Aaron~C
  Courville.
\newblock Improved training of wasserstein gans.
\newblock In {\em Advances in Neural Information Processing Systems}, pages
  5767--5777, 2017.

\bibitem{maskrcnn}
Kaiming He, Georgia Gkioxari, Piotr Doll{\'{a}}r, and Ross~B. Girshick.
\newblock Mask {R-CNN}.
\newblock {\em CoRR}, abs/1703.06870, 2017.

\bibitem{he2016deep}
Kaiming He, Xiangyu Zhang, Shaoqing Ren, and Jian Sun.
\newblock Deep residual learning for image recognition.
\newblock In {\em Proceedings of the IEEE conference on computer vision and
  pattern recognition}, pages 770--778, 2016.

\bibitem{henderson2018learning}
Paul Henderson and Vittorio Ferrari.
\newblock Learning to generate and reconstruct 3d meshes with only 2d
  supervision.
\newblock {\em arXiv preprint arXiv:1807.09259}, 2018.

\bibitem{insafutdinov2018unsupervised}
Eldar Insafutdinov and Alexey Dosovitskiy.
\newblock Unsupervised learning of shape and pose with differentiable point
  clouds.
\newblock In {\em Advances in Neural Information Processing Systems}, pages
  2802--2812, 2018.

\bibitem{isola2017image}
Phillip Isola, Jun-Yan Zhu, Tinghui Zhou, and Alexei~A Efros.
\newblock Image-to-image translation with conditional adversarial networks.
\newblock In {\em Proceedings of the IEEE conference on computer vision and
  pattern recognition}, pages 1125--1134, 2017.

\bibitem{kanazawa2018learning}
Angjoo Kanazawa, Shubham Tulsiani, Alexei~A Efros, and Jitendra Malik.
\newblock Learning category-specific mesh reconstruction from image
  collections.
\newblock In {\em Proceedings of the European Conference on Computer Vision
  (ECCV)}, pages 371--386, 2018.

\bibitem{NMR}
Hiroharu Kato, Yoshitaka Ushiku, and Tatsuya Harada.
\newblock Neural 3d mesh renderer.
\newblock In {\em Proceedings of the IEEE Conference on Computer Vision and
  Pattern Recognition}, pages 3907--3916, 2018.

\bibitem{kingma2014adam}
Diederik~P Kingma and Jimmy Ba.
\newblock Adam: A method for stochastic optimization.
\newblock {\em arXiv preprint arXiv:1412.6980}, 2014.

\bibitem{lambert1760photometria}
J.H. Lambert.
\newblock {\em Photometria}.
\newblock 1760.

\bibitem{li2018differentiable}
Tzu-Mao Li, Miika Aittala, Fr{\'e}do Durand, and Jaakko Lehtinen.
\newblock Differentiable monte carlo ray tracing through edge sampling.
\newblock In {\em SIGGRAPH Asia 2018 Technical Papers}, page 222. ACM, 2018.

\bibitem{liu2018paparazzi}
Hsueh-Ti~Derek Liu, Michael Tao, and Alec Jacobson.
\newblock Paparazzi: Surface editing by way of multi-view image processing.
\newblock In {\em SIGGRAPH Asia 2018 Technical Papers}, page 221. ACM, 2018.

\bibitem{liuadvgeo2018}
Hsueh{-}Ti~Derek Liu, Michael Tao, Chun{-}Liang Li, Derek Nowrouzezahrai, and
  Alec Jacobson.
\newblock Beyond pixel norm-balls: Parametric adversaries using an analytically
  differentiable renderer.
\newblock In {\em ICLR}, 2019.

\bibitem{liu2019soft}
Shichen Liu, Weikai Chen, Tianye Li, and Hao Li.
\newblock Soft rasterizer: Differentiable rendering for unsupervised
  single-view mesh reconstruction.
\newblock {\em arXiv preprint arXiv:1901.05567}, 2019.

\bibitem{liu2019soft_v2}
Shichen Liu, Tianye Li, Weikai Chen, and Hao Li.
\newblock Soft rasterizer: A differentiable renderer for image-based 3d
  reasoning, 2019.

\bibitem{loper2014opendr}
Matthew~M Loper and Michael~J Black.
\newblock Opendr: An approximate differentiable renderer.
\newblock In {\em European Conference on Computer Vision}, pages 154--169.
  Springer, 2014.

\bibitem{luna2012introduction}
Frank Luna.
\newblock {\em Introduction to 3D game programming with DirectX 11}.
\newblock Stylus Publishing, LLC, 2012.

\bibitem{mirza2014conditional}
Mehdi Mirza and Simon Osindero.
\newblock Conditional generative adversarial nets.
\newblock {\em arXiv preprint arXiv:1411.1784}, 2014.

\bibitem{petersen2019pix2vex}
Felix Petersen, Amit~H Bermano, Oliver Deussen, and Daniel Cohen-Or.
\newblock Pix2vex: Image-to-geometry reconstruction using a smooth
  differentiable renderer.
\newblock {\em arXiv preprint arXiv:1903.11149}, 2019.

\bibitem{Phong1975}
Bui~Tuong Phong.
\newblock Illumination for computer generated pictures.
\newblock {\em Commun. ACM}, 18(6):311--317, June 1975.

\bibitem{spheircalharmonic}
Ravi Ramamoorthi and Pat Hanrahan.
\newblock An efficient representation for irradiance environment maps.
\newblock In {\em Proceedings of the 28th Annual Conference on Computer
  Graphics and Interactive Techniques}, SIGGRAPH '01, pages 497--500, New York,
  NY, USA, 2001. ACM.

\bibitem{ronneberger2015u}
Olaf Ronneberger, Philipp Fischer, and Thomas Brox.
\newblock U-net: Convolutional networks for biomedical image segmentation.
\newblock In {\em International Conference on Medical image computing and
  computer-assisted intervention}, pages 234--241. Springer, 2015.

\bibitem{smith19a}
Edward Smith, Scott Fujimoto, Adriana Romero, and David Meger.
\newblock {GEOM}etrics: Exploiting geometric structure for graph-encoded
  objects.
\newblock In {\em Proceedings of the 36th International Conference on Machine
  Learning}, volume~97 of {\em Proceedings of Machine Learning Research}, pages
  5866--5876, Long Beach, California, USA, 09--15 Jun 2019. PMLR.

\bibitem{3DIWGAN}
Edward~J Smith and David Meger.
\newblock Improved adversarial systems for 3d object generation and
  reconstruction.
\newblock In {\em Conference on Robot Learning}, pages 87--96, 2017.

\bibitem{szabo2018unsupervised}
Attila Szab{\'o} and Paolo Favaro.
\newblock Unsupervised 3d shape learning from image collections in the wild.
\newblock {\em arXiv preprint arXiv:1811.10519}, 2018.

\bibitem{tatarchenko2019single}
Maxim Tatarchenko, Stephan~R Richter, Ren{\'e} Ranftl, Zhuwen Li, Vladlen
  Koltun, and Thomas Brox.
\newblock What do single-view 3d reconstruction networks learn?
\newblock In {\em Proceedings of the IEEE Conference on Computer Vision and
  Pattern Recognition}, pages 3405--3414, 2019.

\bibitem{wang2018pixel2mesh}
Nanyang Wang, Yinda Zhang, Zhuwen Li, Yanwei Fu, Wei Liu, and Yu-Gang Jiang.
\newblock Pixel2mesh: Generating 3d mesh models from single rgb images.
\newblock In {\em Proceedings of the European Conference on Computer Vision
  (ECCV)}, pages 52--67, 2018.

\bibitem{wang2018high}
Ting-Chun Wang, Ming-Yu Liu, Jun-Yan Zhu, Andrew Tao, Jan Kautz, and Bryan
  Catanzaro.
\newblock High-resolution image synthesis and semantic manipulation with
  conditional gans.
\newblock In {\em Proceedings of the IEEE Conference on Computer Vision and
  Pattern Recognition}, pages 8798--8807, 2018.

\bibitem{cub}
P.~Welinder, S.~Branson, T.~Mita, C.~Wah, F.~Schroff, S.~Belongie, and
  P.~Perona.
\newblock {Caltech-UCSD Birds 200}.
\newblock Technical Report CNS-TR-2010-001, California Institute of Technology,
  2010.

\bibitem{woo1999opengl}
Mason Woo, Jackie Neider, Tom Davis, and Dave Shreiner.
\newblock {\em OpenGL programming guide: the official guide to learning OpenGL,
  version 1.2}.
\newblock Addison-Wesley Longman Publishing Co., Inc., 1999.

\bibitem{wu2018learning}
Jiajun Wu, Chengkai Zhang, Xiuming Zhang, Zhoutong Zhang, William~T Freeman,
  and Joshua~B Tenenbaum.
\newblock Learning shape priors for single-view 3d completion and
  reconstruction.
\newblock In {\em Proceedings of the European Conference on Computer Vision
  (ECCV)}, pages 673--691. Springer, 2018.

\bibitem{pascal3d}
Yu~Xiang, Roozbeh Mottaghi, and Silvio Savarese.
\newblock Beyond pascal: A benchmark for 3d object detection in the wild.
\newblock In {\em IEEE Winter Conference on Applications of Computer Vision
  (WACV)}, 2014.

\bibitem{yang20173d}
Bo~Yang, Hongkai Wen, Sen Wang, Ronald Clark, Andrew Markham, and Niki Trigoni.
\newblock 3d object reconstruction from a single depth view with adversarial
  learning.
\newblock In {\em Proceedings of the IEEE International Conference on Computer
  Vision}, pages 679--688, 2017.

\bibitem{yao20183d}
Shunyu Yao, Tzu~Ming Hsu, Jun-Yan Zhu, Jiajun Wu, Antonio Torralba, Bill
  Freeman, and Josh Tenenbaum.
\newblock 3d-aware scene manipulation via inverse graphics.
\newblock In {\em Advances in Neural Information Processing Systems}, pages
  1887--1898, 2018.

\end{thebibliography}
\bibliographystyle{plain}
\end{document}